\documentclass[journal]{IEEEtran}

\usepackage{enumerate}
\usepackage{graphicx}
\usepackage{epsf}
\usepackage{subfigure}
\usepackage{amsmath}
\usepackage{amssymb}
\usepackage{array}
\usepackage{setspace}
\usepackage[amsmath,thmmarks]{ntheorem}
\usepackage[ruled,lined]{algorithm2e}
\usepackage{algorithmic}
\usepackage{color}
\usepackage{amsfonts}
\usepackage{dsfont}
\usepackage{xfrac}
\usepackage{breqn}
\usepackage{bbm}
\usepackage{cite}
\usepackage{epstopdf}

\graphicspath{{Fig/},{fig/}}

\theoremseparator{.}

\newcommand{\tabincell}[2]{\begin{tabular}{@{}#1@{}}#2\end{tabular}}

\DeclareMathOperator*{\argmax}{\arg\!\max}

\begin{document}

\title{\LARGE Multi-Functional RIS-Enabled in SAGIN for IoT: A Hybrid Deep Reinforcement Learning Approach with Compressed Twin-Models}


\author{
Li-Hsiang Shen and
Jyun-Jhe Huang
}


\maketitle

\begin{abstract}
A space-air-ground integrated network (SAGIN) for Internet of Things (IoT) network architecture is investigated, empowered by multi-functional reconfigurable intelligent surfaces (MF-RIS) capable of simultaneously reflecting, amplifying, and harvesting wireless energy. The MF-RIS plays a pivotal role in addressing the energy shortages of low-Earth orbit (LEO) satellites operating in the shadowed regions, while accounting for both communication and computing energy consumption across the SAGIN nodes. To maximize the long-term energy efficiency (EE) of IoT devices, we formulate a joint optimization problem over the MF-RIS parameters, including signal amplification, phase-shifts, energy harvesting ratio, and active element selection as well as the SAGIN parameters of beamforming vectors, high-altitude platform station (HAPS) deployment, IoT device association, and computing capability. The formulated problem is highly non-convex and non-linear and contains mixed discrete-continuous parameters. To tackle this, we conceive a compressed hybrid twin-model enhanced multi-agent deep reinforcement learning (CHIMERA) framework, which integrates semantic state-action compression and parametrized sharing under hybrid reinforcement learning to efficiently explore suitable complex actions. The simulation results have demonstrated that the proposed CHIMERA scheme substantially outperforms the conventional benchmarks, including fixed-configuration or non-harvesting MF-RIS, traditional RIS, and no-RIS cases, as well as centralized and multi-agent deep reinforcement learning baselines in terms of the highest EE. Moreover, the proposed SAGIN-MF-RIS architecture in IoT network achieves superior EE performance due to its complementary coverage, offering notable advantages over either standalone satellite, aerial, or ground-only deployments.
\end{abstract}


{\let\thefootnote\relax\footnotetext
{Li-Hsiang Shen and Jyun-Jhe Huang are with the Department of Communication Engineering, National Central University, Taoyuan 320317, Taiwan (email: shen@ncu.edu.tw, kenneth900912@g.ncu.edu.tw). (Corresponding Author: Li-Hsiang Shen)}
}

\section{Introduction}

\subsection{Background and Literature Review}
In the revolutionary era of tele-traffic explosion, global communication technology is rapidly shifting from fifth-generation (5G) to sixth-generation (6G) paradigm, driving ever-growing demands for high-coverage and high-performance Internet of Things (IoT) networks \cite{1, iot1}. Compared to conventional 5G networks, 6G aims to achieve significantly higher data rates, lower latency, and broader network coverage while supporting comprehensive connectivity requirements. Beyond focusing solely on improving communication performance, 6G further integrate advanced technologies such as artificial intelligence (AI), edge computing, and the IoT applications. However, as the number of IoT devices grows explosively, terrestrial base stations (BSs) are gradually becoming overburdened and are unable to meet the future communication demands. To address this bottleneck, the space-air-ground integrated network (SAGIN) has been proposed \cite{2,3, iot2}, combining space, aerial, and terrestrial communication infrastructures into a multi-layered heterogeneous network. Note that the term \textit{ground} in SAGIN indicates the conventional terrestrial networks. This architecture not only ensures global coverage via complementary coverages but also enhances network performance through efficient resource allocation, providing resilient support for the realization of 6G IoT networks.

The low-Earth orbit (LEO) satellites have played an indispensable role in the SAGIN architecture \cite{5,6, iot-leo1}. Thanks to their notable features of global coverage, LEO satellites effectively compensate for coverage gaps in terrestrial BSs and aerial platforms, i.e., drones and high-altitude platform station (HAPS). In remote or disaster areas, where terrestrial communication infrastructure is inaccessible \cite{9}, LEOs or HAPS are capable of delivering stable and reliable communication support. Consequently, when conventional terrestrial communication systems approach overload, SAGIN establishes a crucial foundation for complementing coverage holes and maintaining broader and continuous service. Nevertheless, with aerospace advancements and the rapid expansion of ground devices, LEO satellites face multiple challenges such as limited bandwidth and energy \cite{19,my6, iot-leo2}. Although LEO satellites realize global connectivity and high throughput, they still experience a non-negligible pathloss due to long transmission distances. To overcome these hurdles, LEO satellites in SAGIN must closely cooperate with both aerial and terrestrial BSs, leveraging dynamic network topology adjustments and resilient resource optimizations. In particular, HAPS \cite{iot-hap1, iot-hap2} as key nodes at aerial layer operates at stratospheric altitudes closer to \text{Earth's} surface, offering relatively stable connections \cite{11,12} and dynamic adaptations, thus complementing LEO satellites in lower-latency. Moreover, HAPS can perform as collaborative nodes, distributing satellite tasks with greater granularity to terrestrial stations in order to achieve better interference management and load balancing \cite{13,14}. In \cite{newnew}, the authors have also emphasized the secured SAGIN network along with the sea network, elaborating several issues in global communications, edge computing, monitoring/warning and public safety. This multi-layered heterogeneous integration not only further broadens overall coverage but also effectively enhances communication quality and efficiency.

In addressing the challenges of pathloss, frequent dynamic changes, and channel diversity in LEO satellite communications, reconfigurable intelligent surface (RIS) technology has emerged as a highly promising solution in IoT networks \cite{10, 15, my7, iot-ris1, iot-ris2}. By dynamically adjusting the configuration of RIS elements, one can establish a virtual line-of-sight (LoS) path between the transmitter and receiver, effectively bypassing obstacles and mitigating undesirable reflections as well as fading \cite{15, 16,my4}. Incorporating RIS into LEO introduces additional reflection paths between LEOs and ground devices, thereby reducing pathloss and accommodating the rapid channel variations induced by orbital movement. RIS can refine signal quality through precise phase-shift control, mitigating interference under diverse channel conditions. Upon these advantages, paper in \cite{22} specifically emphasizes improving downlink transmission rates by leveraging RIS to reinforce LEO downlink signals via an augmented virtual LoS, enabling more stable transmissions toward target areas. By orchestrating precise signal allocation and configurations, RIS is capable of boosting communication performance and of providing a scalable scheme for LEO network deployment, thereby laying a solid foundation for the LEO-RIS development \cite{23, iot-ris5}. Despite its great potential, RIS still faces several challenges, such as coverage limitations due to its half space reflection property and dependence on external power sources \cite{add2, add3}. To address these issues, new frameworks have been developed such as simultaneously transmitting and reflecting RIS (STAR-RIS) \cite{add4, iot-ris4}. STAR-RIS is capable of both transmitting and reflecting electromagnetic waves, allowing it to control signals across the entire space. This capability enhances coverage and increases network flexibility. Beyond STAR-RIS, the concept of MF-RIS has also been introduced \cite{17, iot-ris3}. Unlike traditional RIS, which relies on external grid power sources for manipulating signals, MF-RIS can additionally harvest energy from radio frequency (RF) signals, enabling self-sustainable operation while reducing the need of batteries or grid power. This energy harvesting (EH) capability improves the energy efficiency (EE) and self-sustainability of MF-RIS. Additionally, the MF-RIS incorporates active RIS functionality to support signal amplification, enabling dynamic enhancement of signal strength when needed. Leveraging these capabilities, the MF-RIS plays a pivotal role in the SAGIN IoT network by compensating for limited solar energy in LEO satellites, boosting signal strength across both the LEO and HAPS layers, and alternating non-line-of-sight (NLoS) connectivity within the SAGIN architecture.

With the rapid advancement of AI and computing capabilities, the application of AI in IoT has become a main research focus \cite{DRLSAGIN2}, particularly in complex and highly dynamic multi-layered systems such as SAGIN \cite{newnew}. Given the highly dynamic topology and rapidly changing channel conditions in SAGIN, traditional static or analytical optimization methods often struggle to adapt effectively. To overcome these challenges, increasing attention has been directed toward advanced AI-based optimization techniques of deep reinforcement learning (DRL) schemes \cite{newnew, DRL1, iot-ris4, iot-dl1, iot-dl2}. By continuously interacting with the varying environment, DRL can progressively learn the optimal strategies and network configurations, offering greater flexibility and adaptability in scenarios constrained by limited bandwidth, energy, computing capability, and multi-device interference. Recent studies in paper \cite{DRLSAGIN1} have demonstrated that using DRL to replace relay nodes with RIS can significantly enhance EE performance. A doubled dueling deep-Q network (DQN)-based DRL framework was proposed to jointly optimize bandwidth allocation and the three dimensional positioning of aerial RIS. The work in \cite{5-1sag} integrates aerial RIS with satellite-based mobile edge computing, and leverages temporal-enhanced deep deterministic policy gradient (DDPG) and twin-delay-based algorithms to accelerate convergence and reduce system costs. Within this framework, DRL enables real-time analysis of dynamic environments and continuously optimizes network parameters and topology.

\subsection{Challenges and Contributions}
	In this work, we focus on enhancing the EE in SAGIN IoT network architecture by incorporating MF-RIS capabilities. However, the system still faces several challenges, including the rapid variation of complex channel conditions caused by the orbital movement of LEOs, the dynamic deployment of HAPS, and distinct constraints and high-dimensional arguments to be determined, as well as complex coordination across the different layers in SAGIN IoT network. Furthermore, the non-stationary environment and partial observation demand adaptive learning strategies capable of generalizing across time-varying topologies and uncertain environments. To address the issues, we propose a compressed hybrid intelligence framework assisted by twin-models and multi-agent DRL systems. The competitive mechanism improves both learning stability and convergence in complex SAGIN-MF-RIS network. Moreover, parameters might include both continuous and discrete properties which cannot be solved by a standalone model. A shared mechanism is introduced between the two models to enable the exchange of partial state representations and action evaluations, enhancing hidden knowledge transfer between different decision making modules. In addition, the high dimensionality of state and action spaces in DRL can lead to significant low training efficiency, slow convergence and high memory requirement. To mitigate this, semantic compression techniques should be designed to reduce dimensionality while preserving essential features. The main contributions of this paper are elaborated as follows:
\begin{itemize}
    \item We have proposed a novel SAGIN-MF-RIS framework, in which the SAGIN IoT network architecture offers complementary global and local coverage through a three-layer network consisting of LEO satellites, HAPS, and ground BSs. The integration of MF-RIS further extends the signal transmission range and enhances system self-sustainability by leveraging EH capabilities, thereby enabling high EE performance.
    
   \item We have formulated a long-term EE maximization problem, considering both communication and computing capabilities. The optimization variables include the MF-RIS parameters of amplitudes, phase-shifts, EH ratios, and element activation states, transmit antenna beamforming, computing cycles, HAPS deployment as well as IoT device association strategies. The proposed problem is constrained by the limited power consumption, battery capacity, required minimum rate per IoT device, total latency of communication-computing, computing capability, and deployment boundaries.

	\item We propose a compressed hybrid twin-model enhanced multi-agent deep reinforcement learning (CHIMERA) framework to tackle the high-dimensional parameters and mixed discrete-continuous action space: (1) \textbf{Hybrid DRL} framework contains DQN dedicated to discrete decisions (MF-RIS element selection, device association, computing cycles, and HAPS grid-based deployment) and DDPG taking care of continuous variables (transmit beamforming as well as MF-RIS amplitude, phase-shifts, and EH ratio); (2) \textbf{Twin-models} provide a parallel hybrid DRL, preventing policy overfitting and allowing to compete to provide a better action set;
(3) \textbf{Parametrized sharing} mechanism is designed to provide the determined continuous/discrete action outputs as DQN/DDPG's inputs; (4) \textbf{Variational autoencoder (VAE)-based semantic compression} is employed to pre-train three compression models tailored for continuous actions, discrete actions, and state representations. The encoder compresses the state-action inputs fed into the hybrid DRL networks, while the decoder reconstructs the original parameters for accurate policy execution.

\item Simulation results have demonstrated that the proposed SAGIN-MF-RIS architecture outperforms the standalone deployments of LEO satellites, HAPS, or ground BSs, as well as the scenarios of fixed-EH configurations, conventional RIS, and no-RIS scenarios under various system settings. Furthermore, the proposed CHIMERA scheme achieves the highest EE compared to centralized learning approaches such as DQN and DDPG, multi-agent systems, conventional optimization techniques, and heuristic methods. Notably, the VAE-based semantic compression accelerates learning process with moderate performance degradation, enhancing overall system scalability.

\end{itemize}

This paper is organized as follows. Section \ref{sec_sys} introduces the system model of SAGIN-MF-RIS and problem formulation. Section \ref{sec_alg} elaborates the proposed CHIMERA framework. Simulation results are provided in Section \ref{sec_sim}, whereas the conclusion is drawn in Section \ref{sec_con}.

\section{System Model and Problem Formulation}
\label{sec_sys}

\begin{figure}[!t]
\centering
\includegraphics[width=3.3in]{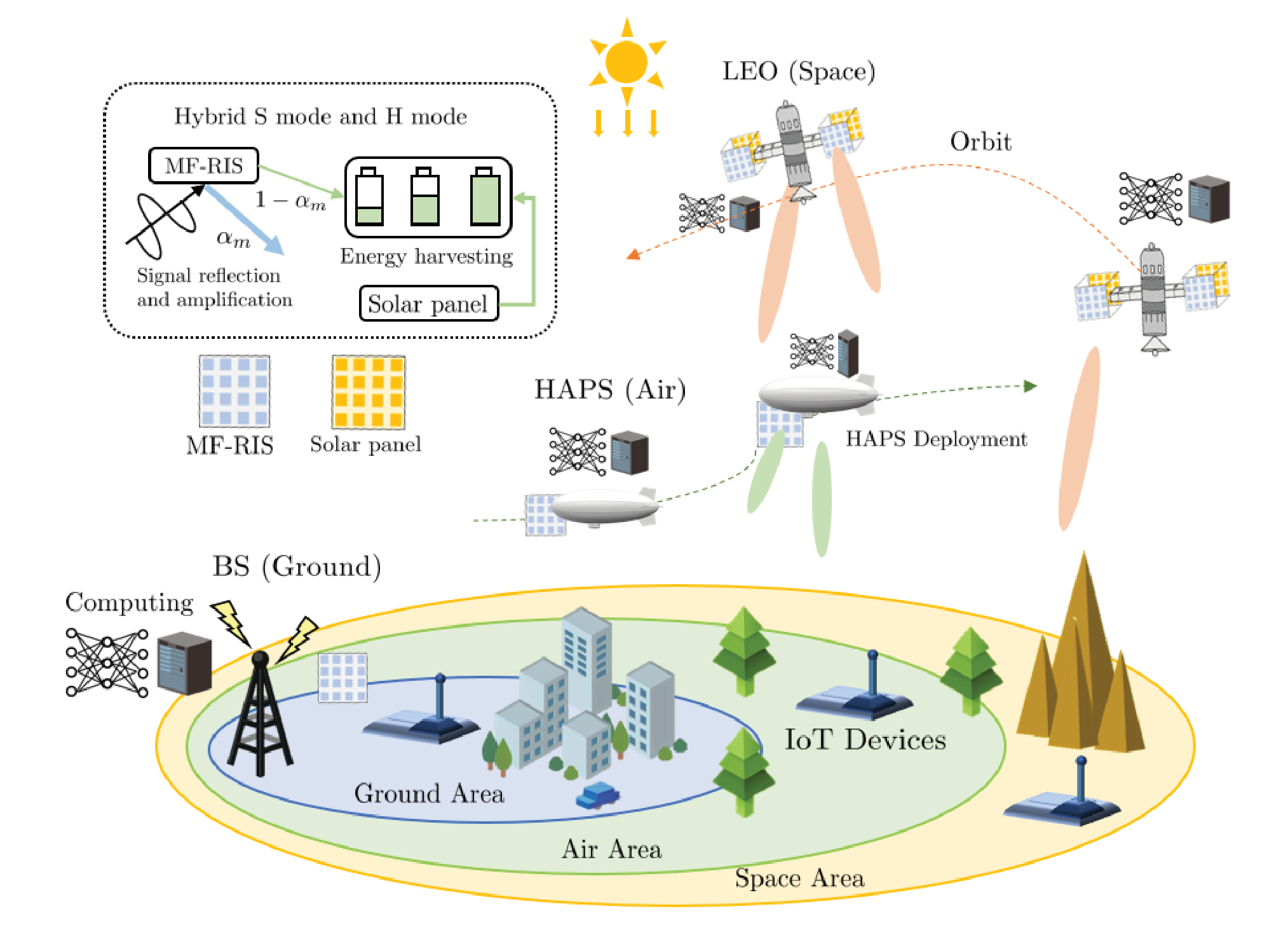}
\caption{The architecture of SAGIN-MF-RIS in IoT network. An MF-RIS is equipped on each SAGIN node, enabling hybrid signal enhancement and energy harvesting.} \label{Fig.1}
\end{figure}
	Fig. \ref{Fig.1} illustrates an MF-RIS-assisted downlink SAGIN IoT network. We consider $N_s$ LEO satellites, $N_a$ HAPS, and $N_g$ BS. These nodes are denoted by the set $\mathcal{N}_c = \{1, 2, \ldots, N_c\}, c\in\mathcal{C}=\{s,a,g\}$, where $N_c$ is the number of transmit nodes at $c$-layer and texts $\{s,a,g\}$ refer to \textit{space}, \textit{air} and \textit{ground}, respectively. Each node is equipped with $N$ transmit antennas, serving total $K$ IoT devices indexed by the set of $ \mathcal{K} = \{1, 2, \ldots, K\}$, with each receiving device equipped with a single antenna. We consider a three-dimensional (3D) Cartesian coordinate system to describe the locations of nodes: The location of LEO/HAPS/BS is represented as $\mathbf{X}_{n_c} = [x_{n_c}, y_{n_c}, h_{n_c}]$, whereas IoT devices are located at $\mathbf{X}_{k} = [x_{k}, y_{k}, h_{k}]$. An MF-RIS with $M$ elements with its set of $\mathcal{M} = \{1, 2, \ldots, M\}$ is equipped on each node for complementing the insufficient power utilization and signal strength. Note that MF-RIS is in a two-dimensional array with a total number of $ M = M_h  \cdot M_v$ elements, where $M_h$ and $M_v$ indicate the respective numbers of elements in horizontal and vertical directions. The MF-RIS configuration on the $n_c$-th node can be defined as $\mathbf{\Theta}_{n_c} \!=\! {\rm diag}\bigg( 
    \alpha_{n_c,1} \sqrt{\beta_{n_c,1}} e^{j\theta_{n_c,1}}, \ldots, \alpha_{n_c,M} \sqrt{\beta_{n_c,M}} e^{j\theta_{n_c,M}} 
\bigg)$,
where $ \beta_{n_c,m} \in [0, \beta_{\text{max}}] $ and $ \theta_{n_c,m} \in [0, 2\pi) $ denote the amplitude and the phase-shift coefficients of MF-RIS, respectively. Note that $\beta_{\text{max}} > 1$ implies the amplified signals, whilst $\beta_{\text{max}}  \leq 1$ means a non-amplified passive RIS. We define $s_{n_c,m}$ as the incident signal received by the $m$-th MF-RIS element at the $n_c$-th node. Each element of the MF-RIS can operate in EH mode (H mode) and signal mode (S mode) by adjusting the EH ratio $\alpha_{n_c,m} \in [0, 1]$. Note that $\alpha_{n_c,m} = 1$ implies that MF-RIS operates in S mode, while $\alpha_{n_c,m} = 0$ is for only H mode. Different from \cite{17}, the considered MF-RIS can operate in a hybrid mode when $0 < \alpha_{n_c,m} < 1$. Therefore, the signals harvested and reflected by the $m$-th MF-RIS element on the $n_c$-th node can be modeled as $y_{n_c,m}^{r} = (1-\alpha_{n_c,m})s_{n_c,m}$ and $ y_{n_c,m}^{h} =  \alpha_{n_c,m} \sqrt{\beta_{n_c,m}} e^{j\theta_{n_c,m}} s_{n_c,m}$, respectively. Moreover, for flexibly utilizing MF-RIS elements, we define $\mathbf{F}_{n_c}={\rm diag} (f_{n_c,1},f_{n_c,2},...,f_{n_c,M})\in\{0,1\}^{M\times M}$ as the element selection matrix with each entity as 0 or 1. Note that $f_{n_c,m}=1$ indicates that the $m$-th element of MF-RIS is under operation, whereas $f_{n_c,m}=0$ indicates an element switched-off.

\subsection{Channel Model}

	We consider the Rician fading channel model \cite{24} between the $n_c$-th node and the $n_c'$-th MF-RIS as $\mathbf{H}_{n_c',n_c} \in \mathbb{C}^{M \times N}$
\begin{align}
	\mathbf{H}_{n_c, n_c'} \!=\! \sqrt{ \frac{h_0}{d_{n_c,n_c'}^{k_0}} }
\left( \sqrt{\frac{\beta_0}{\beta_0 \!+\! 1}} \mathbf{H}_{\text{LoS}} \!+\! \sqrt{\frac{1}{\beta_0 \!+\! 1}} \mathbf{H}_{\text{NLoS}} \right),
\label{4}
\end{align}
where $h_0$ is the pathloss at the reference distance of 1 meter, $d_{n_c,n_c'}$ is the distance, and $k_0$ is the pathloss exponent. Notation of $\beta_0$ is the Rician factor, adjusting the portion of LoS path $\mathbf{H}_{\text{LoS}}$ and NLoS component $\mathbf{H}_{\text{NLoS}}$. The LoS component $ \mathbf{H}_{\text{LoS}}$ is expressed as the array response vector given by
\begingroup
\allowdisplaybreaks
\begin{align}
&\mathbf{H}_{\text{LoS}} 
	= 
\begin{bmatrix}
1,e^{-j\frac{2\pi}{\lambda}d\sin\varphi_r\sin\vartheta_r}, \ldots,e^{-j\frac{2\pi}{\lambda}(M_z-1)d\sin\varphi_r\sin\vartheta_r}
\end{bmatrix}^{\mathcal{T}} \notag \\
	&\otimes
\begin{bmatrix}
1,e^{-j\frac{2\pi}{\lambda}d\sin\varphi_r\cos\vartheta_r}, \ldots,e^{-j\frac{2\pi}{\lambda}(M_y-1)d\sin\varphi_r\cos\vartheta_r}
\end{bmatrix}^{\mathcal{T}} \notag \\
	&\otimes
\begin{bmatrix}
1,e^{-j\frac{2\pi}{\lambda}d\sin\varphi_t\cos\vartheta_t}, \ldots,e^{-j\frac{2\pi}{\lambda}(N-1)d\sin\varphi_t\cos\vartheta_t}
\end{bmatrix},
\label{5}
\end{align}
\endgroup
where $\otimes$ denotes the Kronecker product and $\mathcal{T}$ is transpose operation. Notation of $\lambda$ indicates the wavelength of the operating frequency, and $d$ denotes the antenna and element separation. In \eqref{5}, $\{\varphi_r, \vartheta_r \}$ represents the vertical/horizontal angle-of-arrivals, whilst $\{\varphi_t, \vartheta_t\}$ denotes the vertical/horizontal angle-of-departures. Note that $\mathbf{H}_{\text{NLoS}}$ follows independent and identically distributed Rayleigh fading. The channel matrices from MF-RIS $n_c$ to MF-RIS $n'_{c}$ is denoted as $\mathbf{R}_{n_{c}',n_c} \in \mathbb{C}^{M \times M }, \forall n_{c}'\neq n_{c}$, which follows an asymptotic form in \eqref{4}. Following \eqref{4}, the channel vectors from node $n_c$ to device $k$ and from the MF-RIS $n_c$ to IoT device $k$ are respectively denoted as $\mathbf{h}_{n_c,k} \in \mathbb{C}^{N \times1 }$ and $\mathbf{r}_{n_c,k} \in \mathbb{C}^{M \times 1 }$, where the LoS components are respectively expressed as $\mathbf{h}_{\text{LoS}} = \left[
1, e^{-j\frac{2\pi}{\lambda}d\sin\varphi_t\sin\vartheta_t}, \ldots, e^{-j\frac{2\pi}{\lambda}(N-1)d\sin\varphi_t\sin\vartheta_t}
\right]^{\mathcal{T}}$ and $\mathbf{r}_{\text{LoS}}= \left[
1, e^{-j\frac{2\pi}{\lambda}d\sin\varphi_t\sin\vartheta_t}, \ldots, e^{-j\frac{2\pi}{\lambda}(M-1)d\sin\varphi_t\sin\vartheta_t}
\right]^{\mathcal{T}}$.
For simplicity, we neglect the remaining parameters of $\mathbf{R}_{n_c,n_{c'}}$, $\mathbf{h}_{n_c,k}$, and $\mathbf{r}_{n_c,k}$ due to similar definitions in \eqref{4} and \eqref{5}. We consider cascaded MF-RIS-based channels with up to four signal bounces, accounting for the layered structure of SAGIN. Notably, only signal transmissions to the same or lower layers are feasible; reflections to upper layers are neglected due to the significantly longer propagation distances involved. Then the combined channels from BS/HAPS/LEO node, ground/aerial/space MF-RISs to IoT device $k$ are respectively given by
\begingroup
\allowdisplaybreaks
\begin{subequations} \label{chh}
\begin{align}
	\mathbf{g}_{n_g,k}&  = \underbrace{\mathbf{h}_{n_g,k}^H}_{\text{Direct Link}}  + \sum_{n_g' \in \mathcal{N}_g } \underbrace{\mathbf{r}_{n_g',k}^H \mathbf{F}_{n_g'}\mathbf{\Theta}_{n_g'} \mathbf{H}_{n_g,n_g'}}_{\text{BS-Ground-Device}},
\label{6}
\\
	\mathbf{g}_{n_a,k} & =\mathbf{h}_{n_a,k}^H  + \sum_{n_a' \in \mathcal{N}_a } \underbrace{\mathbf{r}_{n_a',k}^H \mathbf{F}_{n_a'}\mathbf{\Theta}_{n_a'} \mathbf{H}_{n_a,n_a'}}_{\text{HAPS-Aerial-Device}} \notag\\
& + \sum_{n_g' \in \mathcal{N}_g } \underbrace{\mathbf{r}_{n_g',k}^H \mathbf{F}_{n_g'}\mathbf{\Theta}_{n_g'} \mathbf{H}_{n_a,n_a'}}_{\text{HAPS-Ground-Device}} \notag \\
& + \sum_{n_a' \in \mathcal{N}_a}\sum_{n_g' 
\in \mathcal{N}_g} \underbrace{\mathbf{r}_{n_g',k}^H \mathbf{F}_{n_g'}\mathbf{\Theta}_{n_g'} \mathbf{R}_{n_a',n_g'}\mathbf{F}_{n_a'}\mathbf{\Theta}_{n_a'} \mathbf{H}_{n_a,n_a'}}_{\text{HAPS-Aerial-Ground-Device}},
\label{7}
\\
	\mathbf{g}_{n_s,k} & =\mathbf{h}_{n_s,k}^H  + \sum_{c\in\mathcal{C}} \sum_{n_c' \in \mathcal{N}_c } \underbrace{\mathbf{r}_{n_c',k}^H \mathbf{F}_{n_c'}\mathbf{\Theta}_{n_c'} \mathbf{H}_{n_s,n_c'}}_{\text{LEO-Space/Aerial/Ground-Device}}  \notag \\ 
	&+ \sum_{n_s' \in \mathcal{N}_s}\sum_{n_a' 
\in \mathcal{N}_a} \underbrace{\mathbf{r}_{n_a',k}^H \mathbf{F}_{n_a'}\mathbf{\Theta}_{n_a'}\mathbf{R}_{n_s',n_a'}\mathbf{F}_{n_s'}\mathbf{\Theta}_{n_s'} \mathbf{H}_{n_s,n_s'}}_{\text{LEO-Space-Aerial-Device}} \notag \\ 
	&+ \sum_{n_s' \in \mathcal{N}_s}\sum_{n_g' 
\in \mathcal{N}_g} \underbrace{\mathbf{r}_{n_g',k}^H \mathbf{F}_{n_g'}\mathbf{\Theta}_{n_g'}\mathbf{R}_{n_s',n_g'}\mathbf{F}_{n_s'}\mathbf{\Theta}_{n_s'} \mathbf{H}_{n_s,n_s'}}_{\text{LEO-Space-Ground-Device}} \notag \\
	&+ \sum_{n_a' \in \mathcal{N}_a}\sum_{n_g '
\in \mathcal{N}_g} \underbrace{\mathbf{r}_{n_g',k}^H \mathbf{F}_{n_g'}\mathbf{\Theta}_{n_g'}\mathbf{R}_{n_a',n_g'}\mathbf{F}_{n_a'}\mathbf{\Theta}_{n_a'} \mathbf{H}_{n_s,n_a'}}_{\text{LEO-Aerial-Ground-Device}} \notag \\
	&+\sum_{n_s' \in \mathcal{N}_s}\sum_{n_a' 
\in \mathcal{N}_a}\sum_{n_g' \in \mathcal{N}_g} \underbrace{\mathbf{r}_{n_g',k}^H \mathbf{F}_{n_g'}\mathbf{\Theta}_{n_g'} \mathbf{R}_{n_a',n_g'}\mathbf{F}_{n_a'}
\mathbf{\Theta}_{n_a'}} \nonumber \\
	&\qquad\qquad\qquad\qquad\qquad  \underbrace{\mathbf{R}_{n_s',n_a'}\mathbf{F}_{n_s'}\mathbf{\Theta}_{n_s'} \mathbf{H}_{n_s,n_s'}}_{\text{LEO-Space-Aerial-Ground-Device}}.
\label{8}
\end{align}
\end{subequations}
\endgroup
Note that $H$ is the Hermitian operation. We further notice that perfect channel\textsuperscript{\ref{note1}}\footnotetext[1]{Imperfect channel information with estimation errors might occur, including phase errors and additional mutual-coupling effects. The imperfection will largely degrade SAGIN-MF-RIS performances. Some potential solutions are listed as follows: (1) Consider the statistical channel models, robust optimization techniques \cite{m_dro} can be applied to guarantee the worst-case performance within the feasible solution regions. (2) Advanced deep learning architectures, such as transfer/federated learning are designed to deal with erroneous wireless environments \cite{15, newnew}. Both methods should be redesigned as future works. \label{note1}} is considered based on the existing works focused on RIS-induced channel estimation \cite{ch_1, ch_2}. The transmitted signal and the transmit beamforming vector of the $k$-th IoT device served by the $n_c$-th SAGIN node are defined as $x_{n_c,k}$ and $ \mathbf{w}_{n_c,k} \in \mathbb{C}^{N \times 1}$, respectively. Since space-aerial nodes are moving, a re-association process is required to maintain reliable signal quality. Accordingly, we define the association indicator $\delta_{n_c,k}\in\{0,1\}$, where $\delta_{n_c,k}=1$ if IoT device $k$ is connected to node $n_c$ at layer $c$, and vice versa for $\delta_{n_c,k}=0$. The handover overhead is neglected here as it is considerably smaller compared to the minute-scale orbital movement or aerial deployment. Therefore, the received signal of the $k$-th IoT device served by the $n_c$-th node can be given by
\begingroup
\allowdisplaybreaks
\begin{align} \label{r_signal}
	& y_{n_c,k} = \underbrace{\delta_{n_c,k}\mathbf{g}_{n_c,k} \mathbf{w}_{n_c,k} x_{n_c,k}}_{\text{Desired Signal}}
+ \mathbf{g}_{n_c,k} \sum_{k' \in \mathcal{K} \backslash k} \underbrace{\delta_{n_c,k'}\mathbf{w}_{n_c,k'} x_{n_c,k'}}_{\text{Intra-Node Interference}} \notag \\
	& + \sum_{n_c' \in \mathcal{N}_c  \backslash n_c} \sum_{k' \in \mathcal{K} \backslash k}\underbrace{(1-\delta_{n_c,k'})\mathbf{g}_{n_c',k} \mathbf{w}_{n_c',k'} x_{n_c',k'}}_{\text{Intra-Layer Interference}} \notag \\
	& + \sum_{\bar{c} \in \mathcal{C}  \backslash c} \sum_{k' \in \mathcal{K} \backslash k} \underbrace{(1-\delta_{n_c,k'})\mathbf{g}_{n_{\bar{c}},k} \mathbf{w}_{n_{\bar{c}},k'} x_{n_{\bar{c}},k'}}_{\text{Inter-Layer Interference}} + n_{n_c,k},
\end{align}
\endgroup
where $ n_{n_c,k} \sim \mathcal{CN}(0, \sigma_{n_c,k}^2) $ denotes complex additive white Gaussian noise (AWGN) with noise power $ \sigma_{n_c,k}^2$. According to \eqref{r_signal}, the corresponding signal-to-interference-plus-noise ratio (SINR) is given by
\begingroup
\allowdisplaybreaks
\begin{align} \label{sinr}
\gamma_{n_c,k} = 
\frac{
    \left| \delta_{n_c,k} \mathbf{g}_{n_c,k} \mathbf{w}_{n_c,k} \right|^2
}{
    z_1 + z_2 + z_3 + \sigma_{n_c,k}^2
},
\end{align}
\endgroup
where 
$z_1 = 
    \sum_{k' \in \mathcal{K} \backslash k} \left| \delta_{n_c,k'}\mathbf{g}_{n_c,k} \mathbf{w}_{n_c,k'} \right|^2$,
$z_2 = \sum_{n_c' \in \mathcal{N}_c \backslash n_c} 
    \sum_{k' \in \mathcal{K} \backslash k} \left| (1-\delta_{n_c,k'})\mathbf{g}_{n_c',k} \mathbf{w}_{n_c',k'} \right|^2$, and   
$z_3 = \sum_{\bar{c} \in \mathcal{C} \backslash c} 
    \sum_{k' \in \mathcal{K} \backslash k} \left|  (1-\delta_{n_c,k'})\mathbf{g}_{n_{\bar{c}},k} \mathbf{w}_{n_{\bar{c}},k'} \right|^2$. The detailed symbol definition and its corresponding channel components are elaborated in Table \ref{MF_channel}. The ergodic rate of node $n_c$ serving IoT device $k$ is obtained as $R_{n_c,k} = \log_2 (1 + \gamma_{n_c,k})$.

\begin{table*}[t]
\linespread{1.1}
\centering
\footnotesize
  \caption{Symbol Definition and Channel Components in SAGIN-MF-RIS}
 \begin{tabular}{lll}
  \hline
  \textbf{Symbol} & \textbf{Definition}  & \textbf{Channel Components} \\
  \hline\hline
	$\mathbf{g}_{n_s,k}$ & \tabincell{l}{Combined channel from \\ LEO $n_s$ to device $k$} & 
	\tabincell{l}{(1) Direct channel; (2) LEO $\rightarrow$ space/aerial/ground MF-RIS $\rightarrow$ device; \\
	(3) LEO $\rightarrow$ space MF-RIS $\rightarrow$ aerial MF-RIS $\rightarrow$ device; \\
	(4) LEO $\rightarrow$ space MF-RIS $\rightarrow$ ground MF-RIS $\rightarrow$ device; \\
	(5) LEO $\rightarrow$ aerial MF-RIS $\rightarrow$ ground MF-RIS $\rightarrow$ device; \\
	(6) LEO $\rightarrow$ space MF-RIS $\rightarrow$ aerial MF-RIS $\rightarrow$ ground MF-RIS $\rightarrow$ device} \\ \hline
	$\mathbf{g}_{n_a,k}$ & \tabincell{l}{Combined channel from \\ HAPS $n_a$ to device $k$} & 
	\tabincell{l}{(1) Direct channel; (2) HAPS $\rightarrow$ aerial MF-RIS $\rightarrow$ device; \\
	(3) HAPS $\rightarrow$ ground MF-RIS $\rightarrow$ device; \\
	(4) HAPS $\rightarrow$ aerial MF-RIS $\rightarrow$ ground MF-RIS $\rightarrow$ device} 
	\\ \hline
	$\mathbf{g}_{n_g,k}$ & \tabincell{l}{Combined channel from \\ BS $n_g$ to device $k$} & 
	\tabincell{l}{(1) Direct channel; (2) BS $\rightarrow$ ground MF-RIS $\rightarrow$ device} 
	\\ \hline
	$z_1$ & \tabincell{l}{Inter-device Interference} & Come from device $k'$ at the same node $n_c$ \\ \hline
	$z_2$ &  \tabincell{l}{Intra-layer Interference} & Come from device $k'$ at other nodes $n_c$ in the same layer $c$ \\ \hline
	$z_3$ &  \tabincell{l}{Inter-layer Interference} & Come from device $k'$ at other nodes $n_{\bar{c}}$ in the different layers $\bar{c}$ \\ \hline		
	$\mathbf{h}^{\text{eff}}_{n_s}$ & \tabincell{l}{Total beamforming power \\ at space MF-RIS $n_s$} &  
	\tabincell{l}{LEO $\rightarrow$ space MF-RIS}
	\\ \hline
	$\mathbf{h}^{\text{eff}}_{n_a}$ & \tabincell{l}{Total beamforming power 
	\\ at aerial MF-RIS $n_a$} &  
	\tabincell{l}{(1) HAPS $\rightarrow$ aerial MF-RIS; \\
	(2) LEO $\rightarrow$ space MF-RIS $\rightarrow$ aerial MF-RIS; \\
	(3) LEO $\rightarrow$ aerial MF-RIS}
	\\ \hline
	$\mathbf{h}^{\text{eff}}_{n_g}$ & \tabincell{l}{Total beamforming power \\ at ground MF-RIS $n_g$} &  
	\tabincell{l}{(1) BS $\rightarrow$ ground MF-RIS; \\
	(2) HAPS $\rightarrow$ aerial MF-RIS $\rightarrow$ ground MF-RIS; \\
	(3) HAPS $\rightarrow$ ground MF-RIS; \\
	(4) LEO $\rightarrow$ space MF-RIS $\rightarrow$ aerial MF-RIS $\rightarrow$ ground MF-RIS; \\
	(5) LEO $\rightarrow$ space MF-RIS $\rightarrow$ ground MF-RIS; \\
	(6) LEO $\rightarrow$ aerial MF-RIS $\rightarrow$ ground MF-RIS; \\
	(7) LEO $\rightarrow$ ground MF-RIS;}
	\\						
  \hline
 \end{tabular} \label{MF_channel}
\end{table*}

\subsection{Computing Model} 

The computing capability of SAGIN node as edge is crucial as it shares the same resource pools with communication entity. We define $D^{\text{comm}}_{n_c,k}$ as the size of communication data of IoT device $k$ to be processed and $ U_{n_c}$ as a computing capability in cycles of the node $n_c$. Coherently, the computing latency can be modeled as
\begin{align} \label{del_1}
t^{\text {cp}}_{n_c} = \frac{\sum_{k\in \mathcal{K}} \delta_{n_c,k} D^{\text{comm}}_{n_c,k}}{U_{n_c}}.
\end{align}
The corresponding computing energy cost can be
written as $E^{\text{cp}}_{n_c}=\tau t^{\text {cp}}_{n_c} U_{n_c}^3$, where $\tau$ is a hardware-related constant. Therefore, the averaged data transmission delay of the node $n_c$ is denoted as
\begin{align} \label{del_2}
t^{\text {tr}}_{n_c} = \frac{1}{K_{n_c}} \sum_{k\in\mathcal{K}} \frac{\delta_{n_c,k} D^{\text{comm}}_{n_c,k}}{R_{n_c,k}}, \ \text{ if } R_{n_c,k}\neq 0,
\end{align}
where $K_{n_c} = \sum_{k\in\mathcal{K}} \delta_{n_c,k}$ indicates the number of IoT devices served by node $n_c$. Also, the over-the-air propagation delay between the node and IoT devices is acquired as
\begin{align} \label{del_3}
t^{\text{pr}}_{n_c}=\frac{ \frac{1}{K_{n_c}} \sum_{k\in\mathcal{K}} \delta_{n_c,k} d_{n_c,k}}{c_{sl}},
\end{align}
where $c_{sl}$ is the speed of light and $d_{n_c,k}$ is the distance between node $n_c$ and IoT device $k$. Combining \eqref{del_1}, \eqref{del_2} and \eqref{del_3} yields the total averaged end-to-end latency of node $n_c$ as $t^{\text{tot}}_{n_c}=t^{\text {cp}}_{n_c}+t^{\text {tr}}_{n_c}+t^{\text{pr}}_{n_c}$.

\subsection{Power Dissipation Model}
\subsubsection{MF-RIS Power Model} 
	We define the EH ratio matrix for the $ m $-th element of the $n_c$-th MF-RIS as $\mathbf{T}_{n_c,m} = {\rm diag}\left([ 0, \ldots, 0, 1 - \alpha_{n_c,m}, 0, \ldots, 0 ]\right)$. Accordingly, the RF power received at the $m$-th element of the $n_c$-th MF-RIS is obtained as
\begin{align}
	P_{n_c,m}^{\text{RF}} = \left\lVert  \mathbf{F}_{n_c}\mathbf{T}_{n_c,m} \left( \mathbf{h}^{\text{eff}}_{n_c} + \mathbf{n}_{m} \right) \right\rVert^2,
\end{align}
where $\mathbf{h}_{n_c}^{\text{eff}}$ indicates the effective total channel associated with the pertinent parameters defined in \eqref{RFFF} shown at top of next page. Note that the detailed elaboration of components is shown in Table \ref{MF_channel}.
\begin{figure*}
\begin{subequations} \label{RFFF}
\begin{align}
	\mathbf{h}^{\text{eff}}_{n_s} & = \sum_{n_s' \in \mathcal{N}_s} \sum_{k\in \mathcal{K}} \mathbf{H}_{n_s',n_s} \delta_{n_s',k} \mathbf{w}_{n_s',k}, \label{RF1}
\\
	\mathbf{h}^{\text{eff}}_{n_a} & = \sum_{n_a' \in \mathcal{N}_a} \sum_{k \in \mathcal{K}} \mathbf{H}_{n_a',n_a} \delta_{n_a',k} \mathbf{w}_{n_a',k} + \sum_{\substack{\{n_s,n_s'\} \\ \in \mathcal{N}_s}} \sum_{k \in \mathcal{K}} (\mathbf{R}_{n_s,n_a}\mathbf{F}_{n_s}\mathbf{\Theta}_{n_s}\mathbf{H}_{n_s',n_s} \!+\! \mathbf{H}_{n_s',n_a}) \delta_{n_s',k} \mathbf{w}_{n_s',k}, \label{RF2}
\\
	\mathbf{h}^{\text{eff}}_{n_g} & = \sum_{n_g' \in \mathcal{N}_g} \sum_{k \in \mathcal{K}} \mathbf{H}_{n_g',n_g} \delta_{n_g',k} \mathbf{w}_{n_g',k} 
	+ \sum_{\substack{\{n_a,n_a'\} \\ \in \mathcal{N}_a}} \sum_{k \in \mathcal{K}} (\mathbf{R}_{n_a,n_g}\mathbf{F}_{n_a}\mathbf{\Theta}_{n_a}\mathbf{H}_{n_a',n_a} + \mathbf{H}_{n_a',n_g}) \delta_{n_a',k} \mathbf{w}_{n_a',k} \nonumber \\
	& + \sum_{\substack{\{n_s,n_s'\} \\ \in \mathcal{N}_s}} \sum_{\substack{n_a \in \mathcal{N}_a}} \sum_{k \in \mathcal{K}} \left( \mathbf{R}_{n_a,n_g}\mathbf{F}_{n_a}\mathbf{\Theta}_{n_a}  \mathbf{R}_{n_s,n_a}\mathbf{F}_{n_s}\mathbf{\Theta}_{n_s}\mathbf{H}_{n_s',n_s} 
	+  \mathbf{R}_{n_s,n_g}\mathbf{F}_{n_s}\mathbf{\Theta}_{n_s}\mathbf{H}_{n_s',n_s} 
	\right. \notag \\
	&\left. \qquad\qquad\qquad\qquad\qquad\qquad\qquad\qquad	\qquad\qquad\qquad
	+  \mathbf{R}_{n_a,n_g}\mathbf{F}_{n_a}\mathbf{\Theta}_{n_a}\mathbf{H}_{n_s',n_a}
	+ \mathbf{H}_{n_s',n_g} 
	\right) \delta_{n_s',k} \mathbf{w}_{n_s',k}. \label{RF3}
\end{align}
\end{subequations}
\hrulefill
\end{figure*}
In \eqref{RFFF}, notation of $\mathbf{n}_{m} \sim \mathcal{CN}(0, \sigma_{m}^2 \mathbf{I}_M) $ denotes the amplified noise introduced by the MF-RIS associated with its power $\sigma_m^2$. In order to capture the dynamics of the RF energy conversion efficiency for different input power levels, a non-linear energy harvesting model is employed \cite{17}. Accordingly, the total power harvested from the $m$-th element of the $n_c$-th MF-RIS is given by $P_{n_c,m}^h = \frac{\Upsilon_{n_c,m} - Z \Omega}{1 - \Omega}$ if $P_{n_c,m}^{\text{RF}} \geq P_{\text{th}}^{\text{RF}}$ and $P_{n_c,m}^h=0$ if $P_{n_c,m}^{\text{RF}} < P_{\text{th}}^{\text{RF}}$, where $P_{\text{th}}^{\text{RF}}$ indicates the sensitivity threshold owing to the non-linearity in the low input power region. Notation $ \Upsilon_{n_c,m} = \frac{Z}{1 + e^{-a (P_{n_c,m}^{\text{RF}} - q)}} $ is a logistic function associated with the received RF power $ P_{n_c,m}^{\text{RF}} $, and $ Z \geq 0 $ is a constant determining the maximum harvested power. The constant $\Omega = \frac{1}{1 + e^{a q}}$ is introduced to ensure a zero-input/zero-output response in H mode. Note that the constants $ a > 0 $ and $ q > 0 $ capture the joint effects of circuit sensitivity and current leakage \cite{17}.

Moreover, the power consumed for controlling MF-RIS mainly comes from the total number of PIN diodes required \cite{thzris}. The total number of required PIN diodes can be obtained as $\log_2 L_{\alpha} + \log_2 L_{\beta} +  \log_2 L_{\theta}$, where $L_{\alpha}$, $L_{\beta}$, and $L_{\theta}$ indicate the quantization levels controlled by PIN diodes for the EH coefficient, amplitude, and phase-shifts, respectively. Then, the power consumption of the MF-RIS can be obtained as
\begin{align} \label{Rcon}
P_{n_c,R}^{\text{cons}} & = \frac{1}{2} \left[ \log_2 L_{\alpha} + \log_2 L_{\beta} + \log_2 L_{\theta} \right] \cdot \sum_{m\in\mathcal{M}} f_{n_c,m} \cdot P_{\text{pin}} \nonumber \\
	& \qquad + P_C + \xi \cdot P_{n_c,O},
\end{align}
where $P_{\text{pin}}$ is the power consumption per PIN entity, $P_C$ is the circuital power consumption of the RF-to-DC power conversion, and $\xi$ is the inverse of the power amplifier efficiency. In \eqref{Rcon}, the output power $P_{n_c,O}$ of MF-RIS $n_c$ can be expressed as $P_{n_c,O} = \left\| \mathbf{F}_{n_c}\mathbf{\Theta}_{n_c} \mathbf{h}_{n_c}^{\text{eff}} \right\|^2 + \sum_{m\in \mathcal{M}}\sigma_m^2 \left\| \mathbf{F}_{n_c} \mathbf{\Theta}_{n_c} \right\|_F^2$, where $ \|\cdot\|_F $ is the Frobenius norm.

\begin{figure}[!t]
\centering
\includegraphics[width=3in]{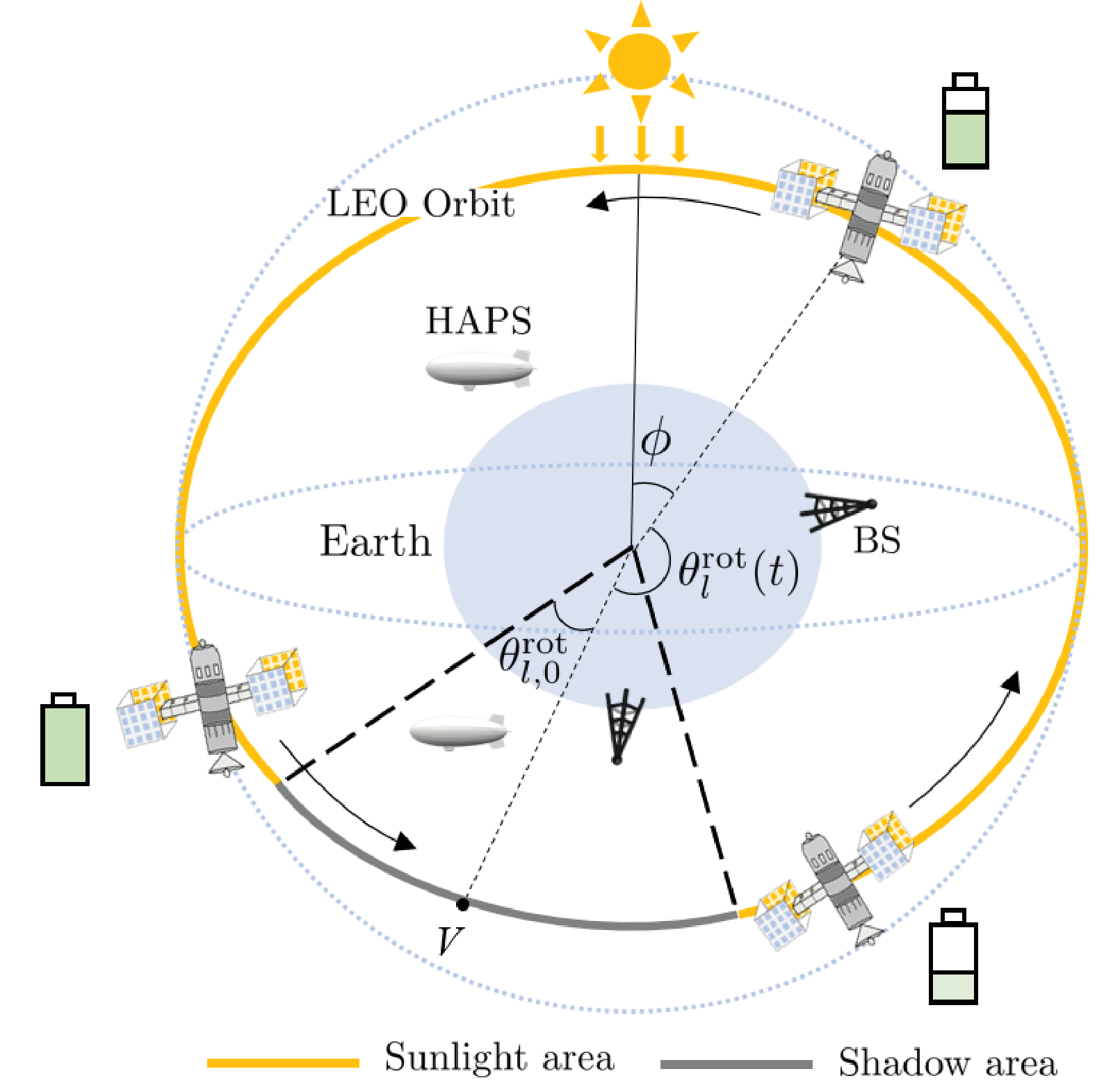}
\caption{LEO satellite's energy consumption when travelling through sunlight and shadow areas.} \label{Fig.sys.3}
\end{figure}

\subsubsection{SAGIN Energy Model} 
We partition the total elapsed time $T$ into multiple time slots, with the interval of each time slot defined as $\Delta$. In time $t$, the energy harvested by the satellite solar panels can be expressed
as \cite{27}
\begin{align} \label{solar}
	E_{n_s}^{\text{sol}}(t) = \int_t^{t+\Delta} \eta_s \psi B \sqrt{1 - \cos^2\phi \cos^2\theta_{n_s}^{\text{rot}}(t)} \, {\rm d}t,
\end{align}
where $\eta_s$ is the efficiency of energy conversion, $\psi$ is the light intensity, and $B$ is the size of the solar panels. The notation of $\phi$ indicates the angle between the satellite orbital plane and the sunlight direction, whereas $\theta_{n_s}^{\text{rot}}(t)$ defines the LEO rotation angle from the midpoint of the shaded orbit $V$ at time $t$. Fig. \ref{Fig.sys.3} illustrates an example of the LEO satellite's energy consumption when travelling through the sunlight and shadow regions. In sunlight area, the battery is charged by harvesting the energy from both the solar panels and signal energy from the MF-RIS. However, in the shadow region where solar energy is unavailable, LEO relies primarily on the remained battery capacity and the extra energy harvested from the MF-RIS. We define the half-angle of the shadow area that the LEO satellite travels as \cite{27}
\begin{align}
\theta_{n_s,0}^{\text{rot}} = 
\begin{cases} 
0, \quad \text{ if } \phi > \sin^{-1} \left( \frac{R_e}{R_e + h_{n_s}} \right) ,\\
\sin^{-1} \left( \frac{R_e^2 \cos^2\phi - \left(2 R_e h_{n_s} + h_{n_s}^2 \right) \sin^2\phi}{(R_e + h_{n_s}) \cos \phi} \right), & \text{otherwise}, 
\end{cases}
\end{align}
where $R_e$ is the radius of the Earth and $h_{n_s}$ is the satellite's height. Note that the range of  $\theta_{n_s,0}^{\text{rot}}$ is $(- \pi, \pi)$. When the satellite is in the sunlight area of $|\theta_{n_s}^{\text{rot}}(t)| \geq \theta_{n_s,0}^{\text{rot}}$, the remaining time it takes to shift from its current position $\theta_{n_s}^{\text{rot}}(t)$ to the shadow area can be expressed as
\begin{align}
	T_{n_s}^{\text{sun}}(t) = 
\begin{cases} 
\frac{2\pi - \theta_{n_s,0}^{\text{rot}} - \theta_{n_s}^{\text{rot}}(t)}{\dot{\Omega}}, & \text{ if }\theta_{n_s}^{\text{rot}}(t) \in [0, \pi), \\
\frac{-\theta_{n_s,0}^{\text{rot}} - \theta_{n_s}^{\text{rot}}(t)}{\dot{\Omega}}, & \text{ if }\theta_{n_s}^{\text{rot}}(t) \in [-\pi, 0) ,
\end{cases}
\end{align}
where $\dot{\Omega}$ is the Earth rotation rate. Moreover, the duration of LEO $n_s$ moving from the current location in shadow area to the sunlight area can be expressed as $T_{n_s}^{\text{shd}}(t) = \frac{\theta_{n_s,0}^{\text{rot}} - \theta_{n_s}^{\text{rot}}(t)}{\dot{\Omega}}$. Then the charging solar power of LEO battery can be denoted as $P_{n_s,\text{in}}^b(t) = \frac{E_{n_s}^{\text{sun}}(t)}{T_{n_s}^{\text{sun}}(t)}$. When the LEO satellite is in the sunlight area, the remaining battery energy on LEO $n_s$ at time  $t$ can be expressed as
\begingroup
\allowdisplaybreaks
\begin{align}
& E_{n_s}(t) \!=\! 
	\min\left( E^b,  E_{n_s}(t-1) + \left[ P_{n_s,\text{in}}^b(t)\cdot \mathbbm{1}(\mathcal{V}=\text{sun}) \right.\right. \notag \\
	& \left.\left. +\sum_{m \in \mathcal{M}} P_{n_s,m}^h - P_{n_s,R}^{\text{cons}} - (P_{n_s}^{\text{tr}} \!+\! P_{n_s}^{\text{cons}}) \right] \cdot T_{n_s}(t\!-\!1) \right), \label{19}
\end{align}
\endgroup
where $\mathcal{V}\in \{ \text{sun}, \text{shd}\}$ indicates whether the current LEO location is in sunlight or shadow area, and $\mathbbm{1}(\cdot)$ indicates the event occurrence. In \eqref{19}, $E^b$ denotes the LEO satellite battery capacity, $P^{\text{tr}}_{n_c} = \sum_{k \in \mathcal{K}} \|\mathbf{w}_{n_c,k}\|^2$ indicates the transmit power of node $n_c$, and $P_{n_c}^{\text{cons}}$ represents the regular operational circuit power. Different from \eqref{19}, both the HAPS and ground BS adopt a stationary solar energy harvesting model, i.e., $E^{\text{sol}}_{n_c}(t) \approx \eta_{c} \psi B, \forall c\in\{a,g\}$ in \eqref{solar} due to their significantly slower mobility compared to LEO satellites. Moreover, the harvesting amount highly depend on their locations. The pertinent models and settings can be found in \cite{SOLAR, SOLAR2}, which is a simplified version of \eqref{19}. Accordingly, the total energy consumption can be expressed as $E_{n_c}^{\text{cons}}(t) = \left( P_{n_c,R}^{\text{cons}} + P_{n_c}^{\text{tr}} + P_{n_c}^{\text{cons}}
\!-\! \sum_{m \in \mathcal{M}}P_{n_c,m}^h \right) T_{n_c}(t)$,
where $T_{n_s}(t) = T_{n_s}^{\text{sun}}(t) + T_{n_s}^{\text{shd}}(t) = T_{n_a}(t) = T_{n_g}(t)$ represents the total duration for the LEO satellite to complete a full orbit, while the HAPS and BS nodes adopt the same value for fair comparison. Accordingly, the total system energy consumption can be expressed as $E_{n_c}^{\text{tot}}(t) = E_{n_c}^{\text{cons}}(t)+E_{n_c}^{\text{cp}}(t)$. Moreover, since continuous time $t$ cannot be directly evaluated in simulations, we discretize the entire LEO orbital trajectory into a sequence of time steps ${t_1, t_2, \ldots, T}$, each representing a small quantized interval.


\subsection{Problem Formulation}
The objective is to maximize the long-term EE\textsuperscript{\ref{note2}}\footnotetext[2]{The short-term EE can be reformulated as an alternative problem once instantaneous measurements at the millisecond scale are available, in contrast to the fast LEO traversals. In this case, the objective does not involve the long-term averaging term $\lim_{T \to \infty} \frac{1}{T} \sum_{t=0}^{T}$. Accordingly, conventional convex optimization techniques, such as fractional programming, block coordinate descent, Lagrangian dual transformation, successive convex approximation, and integer relaxation can be redesigned in a centralized manner \cite{17, my4, my1}. However, further low-complexity architectural designs are essential to enable decentralized SAGIN solutions for this dynamic and complex convex problems.\label{note2}} while guaranteeing the constraints of minimum IoT device rate requirement, MF-RIS configurations and operational power limitation, which is formulated as follows:
\begingroup
\allowdisplaybreaks
\begin{subequations} \label{prob}
	\begin{align}
& \max_{\substack{\theta_{n_c,m}, \alpha_{n_c,m},  \beta_{n_c,m},\\ 
        \mathbf{w} _{n_c,k}, U_{n_c},\delta_{n_c,k}, \\
        \mathbf{F}_{n_c}, \mathbf{X}_{n_a}}}  \ \lim_{T \to \infty} \frac{1}{T} \sum_{t=0}^{T}\sum_{c \in \mathcal{C}} \sum_{n_c\in 
\mathcal{N}_c} \sum_{k\in \mathcal{K}} \frac{R_{n_c,k}(t)}{E_{n_c}^{\text{tot}}(t)} \label{prob_29} \\
	& \text{s.t.} \quad \mathbf{\Theta}_{n_c} \in \mathcal{R}_{\mathbf{\Theta}}, \quad \forall n_c \in \mathcal{N}_c, \forall c\in\mathcal{C}, \label{29b}\\
	& \qquad R_{n_c,k} \geq R^{\text{min}}_{n_c,k}, \quad \forall n_c \in \mathcal{N}_c, \forall c\in\mathcal{C}, \forall k \in \mathcal{K}, \label{29c} \\
	& \qquad P^{\text{cons}}_{n_c,R} \leq \sum_{m \in \mathcal{M}} P^{h}_{n_c,m}, \quad \forall n_c \in \mathcal{N}_c, \forall c\in\mathcal{C}, \label{29d} \\
	& \qquad \sum_{k \in \mathcal{K}} \|\mathbf{w}_{n_c,k}\|^2  \leq P^{\text{max}}_{n_c}, \quad \forall n_c \in \mathcal{N}_c, \forall c\in\mathcal{C}, \label{29e} \\
	& \qquad E_{n_c}(t) \geq 0, \quad \forall n_c \in \mathcal{N}_c, \forall c\in\mathcal{C} \label{29f},
\\
   & \qquad t^{\text{tot}}_{n_c} \leq t^{\text{th}}_{n_c}, \quad \forall n_c \in \mathcal{N}_c,\forall c\in\mathcal{C}, \label{29g}
\\
  & \qquad \sum_{c\in\mathcal{C}} \sum_{n_c\in\mathcal{N}_c}\delta_{n_c,k}= \{0, 1\}, \quad \forall k \in \mathcal{K}, \label{29h}\\
  & \qquad U_{n_c} \leq U_{\text{max}}, \quad \forall n_c\in\mathcal{N}_{c}, \forall c\in\mathcal{C} \label{29i},\\
  & \qquad f_{n_c} \in \{0,1\}, \quad \forall n_c\in\mathcal{N}_{c}, \forall c\in\mathcal{C} \label{29j},\\
  & \qquad \mathbf{X}_{\text{min}} \preceq \mathbf{X}_{n_a} \preceq \mathbf{X}_{\text{max}}, \quad \forall n_a \in\mathcal{N}_a. \label{29k}
	\end{align}
\end{subequations}
\endgroup
Constraints \eqref{29b} defines the constraint set of MF-RIS coefficients as $\mathcal{R}_{\mathbf{\Theta}}$, including $\alpha_{n_c,m} \in [0,1]$, $\beta_{n_c,m} \in [0, \beta_{\text{max}}]$, and $\theta_{n_c,m} \in [0, 2\pi)$. Constraint \eqref{29c} guarantees the minimum rate requirement of each IoT device as $R_{n_c,k}^{\text{min}}$. Constraint \eqref{29d} ensures the self-sustainability of MF-RIS, i.e., power consumption of the MF-RIS cannot exceed its harvested power. Constraint \eqref{29e} guarantees that transmit power should be smaller than $P^{\text{max}}_{n_c}$. Constraint \eqref{29f} ensures that the remaining battery energy must be greater than zero. Constraint \eqref{29g} ensures that the total latency must not exceed $t^{\text{th}}_{n_c}$. Constraint \eqref{29h} ensures that each IoT device is served by only one node. Constraint \eqref{29i} confines its computing upper bound of $U_{\text{max}}$, whereas \eqref{29j} restricts binary element selection variable. The last constraint \eqref{29k} limits the HAPS deployment boundary within $\mathbf{X}_{\text{min}}$ and $\mathbf{X}_{\text{max}}$. Due to the problem's non-convexity, non-linearity, and the mixed high-dimensional discrete and continuous variables under highly-dynamic environment of SAGIN, solving \eqref{prob_29} is compellingly challenging. Hence, we propose a DRL-based approach as detailed in the following section.

\section{Proposed CHIMERA Framework}
\label{sec_alg}

We have conceived a CHIMERA framework for solving problem \eqref{prob}, which consists of a multi-agent architecture, where each node acts as an agent equipped with two pairs of DDPG and DQN to determine its own policy consisting of a plethora of continuous and discrete variables. Furthermore, bi-directional parametrized sharing is adopted to exchange the hidden information, whereas compression technique is designed to significantly reduce parameter dimensions, thereby accelerating the overall learning speed.

\begin{figure}[!t]
\centering
\includegraphics[width=3.3in]{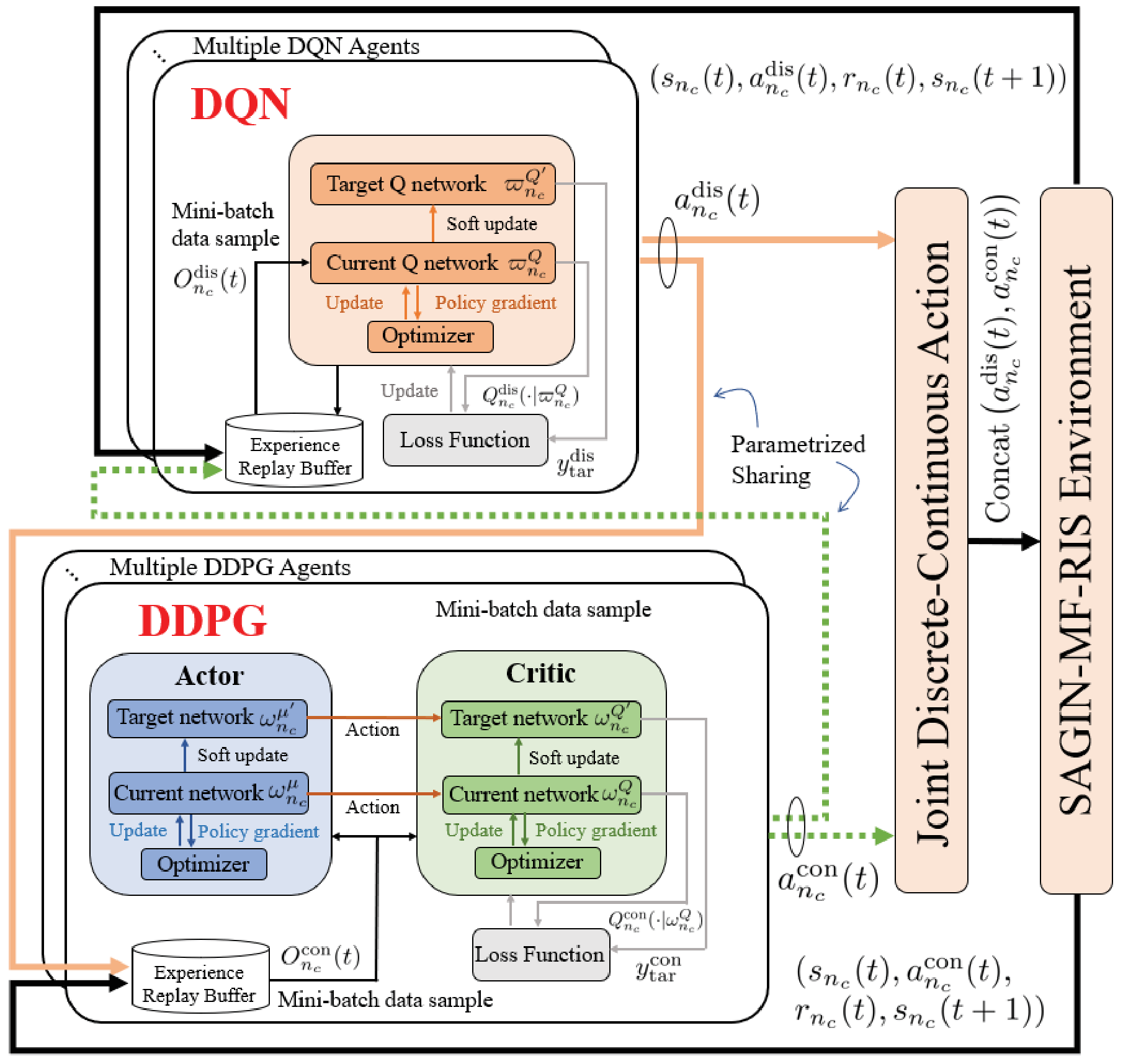}
\caption{The proposed CHIMERA with multi-agent hybrid DRL framework with bi-directional parametrized sharing.}
\label{MAS-H}
\end{figure}

\subsection{Multi-Agent Hybrid DRL}

We consider a typical DRL framework consisting of a state set $\mathcal{S}$, an action set $\mathcal{A}$, and a reward set $\mathcal{R}$. However, a single agent in conventional DRL potentially faces several challenges, including slow convergence, high memory requirement, state collection overhead, and signalling of action from the central server to each node. To address this, we utilize a multi-agent system to reduce the training burden and improve the decision-making efficiency, as depicted in Fig. \ref{MAS-H}. In this framework, an agent representing one node of SAGIN-MF-RIS interacts with a dynamic environment by taking \textit{actions}, receiving the corresponding \textit{rewards}, and thereby updating their \textit{states}, which are defined as follows:
\begin{itemize}
    \item \textbf{State:} The total state space is defined as a set of individual agent state $\mathcal{S} = \{\mathcal{S}_1, \mathcal{S}_2, \ldots, \mathcal{S}_{N_c} | \forall c \in\mathcal{C} \}$. The state $\mathcal{S}_{n_c}$ of each node is composed of a sequence of states over time, denoted as $\mathcal{S}_{n_c} = \{s_{n_c}(1), s_{n_c}(2), \ldots, s_{n_c}(T)\}$, where $s_{n_c}(t) = \{ \mathbf{g}_{{n_c},k}(t) | \forall k \in \mathcal{K} \}$ denotes the combined MF-RIS channel from node $n_c$ to IoT device $k$ at time $t$ in \eqref{7}. To prevent overhead of excessive channel information exchange, we consider that only the channel related to the node itself can be measured.

    \item \textbf{Action:} The total action set is divided into continuous and discrete parts denoted as $ \mathcal{A}^v = \{\mathcal{A}^v_{1}, \mathcal{A}^v_{2}, \ldots, \mathcal{A}^v_{N_c} | \forall c\in\mathcal{C}\}, \forall v \in \{\text{con},\text{dis}\}$. Note that "con/dis" refers to continuous/discrete actions. Each agent's action set $ \mathcal{A}^v_{n_c}$ consists of a sequence of actions over time, defined as $ \mathcal{A}^v_{n_c}= \{a^v_{n_c}(1), a^v_{n_c}(2), \ldots, a^v_{n_c}(T)\} $. Note that the MF-RIS configurations of amplitude/phase/EH ratio $\{\beta_{{n_c},m}(t), \theta_{{n_c},m}(t), \alpha_{{n_c},m}(t)\}$ and transmit beamforming $\mathbf{w}_{{n_c},k}(t)$ are with the continuous property. While, computing cycle $U_{n_c}(t)$, IoT device association $\delta_{n_c,k}(t)$, MF-RIS element selection $\mathbf{F}_{n_c}(t)$ and HAPS grid-based deployment location $\mathbf{X}_{n_c}(t)$ are regarded as discrete variables. Therefore, the actions $a^v_{n_c}(t)$ of node $n_c$ at time $t$ associated with the parameters to be determined in problem \eqref{prob} are denoted as  $ a^{\text{con}}_{n_c}(t) = \{\beta_{{n_c},m}(t), \theta_{{n_c},m}(t), \alpha_{{n_c},m}(t), \mathbf{w}_{{n_c},k}(t) | \forall m\in \mathcal{M}, \forall k\in \mathcal{K}\} $ and $ a^{\text{dis}}_{n_c}(t) = \{U_{n_c}(t) \delta_{n_c, k}(t), \mathbf{F}_{n_c}(t),\mathbf{X}_{n_a}(t)| \forall m\in \mathcal{M}, \forall k\in \mathcal{K}\} $. Note that $\mathbf{X}_{n_g}(t)$ is fixed as the ground infrastructure, whereas $\mathbf{X}_{n_s}(t)$ is automatically determined by its orbital movement.

    \item \textbf{Reward:} The reward set is defined as $\mathcal{R} = \{\mathcal{R}_{i}, \mathcal{R}_2, \ldots, \mathcal{R}_{N_c}| \forall c\in\mathcal{C} \}$, where the individual reward set is given by $\mathcal{R}_{n_c} = \{r_{n_c}(1), r_{n_c}(2), \ldots, r_{n_c}(T)\}$. Note that the reward $r_{n_c}(t)$ at time $t$ is acquired after executing both discrete and continuous actions, which is designed as a penalized EE as
\begin{align}
	r_{n_c}(t) = \frac{\sum_{k\in \mathcal{K}} R_{n_c,k}(t)}{E_{n_c}^{\text{tot}}(t)} - \sum_{i=1}^{5} \rho_{n_c,i} C_{n_c,i},
\end{align}
where $\rho_{n_c,i} \geq 0$ indicates the weights of each constraint penalty $C_{n_c,i}$. The penalty terms $C_{n_c,i}$ corresponding to the constraints in problem \eqref{prob_29} are designed as $ C_{n_c,1} = \sum_{k\in\mathcal{K}} R_{n_c,k}^{\text{min}} - R_{n_c,k} $, $ C_{n_c,2} = P_{n_c,R}^{\text{cons}}-\sum_{m \in \mathcal{M}} P^{h}_{n_c,m}$, $ C_{n_c,3} = \sum_{k \in \mathcal{K}} \|\mathbf{w}_{n_c,k}\|^2 - P^{\text{max}}_{n_c} $, $ C_{n_c,4} = - E_{n_c}(t)$, and $ C_{n_c,5} = t^{\text{tot}}_{n_c} -  t^{\text{th}}_{n_c}$. It is worth noting that the remaining constraints are inherently satisfied during the action generation process.
\end{itemize}

As shown in Fig. \ref{MAS-H}, the proposed framework integrates multi-agent DQN (MADQN) and multi-agent DDPG (MADDPG) to handle discrete and continuous actions, respectively. Note that all DDPG and DQN networks share the same state $\mathcal{s}_{n_c}(t)$ as input, ensuring consistent observations and coordinated learning information. Both neural network models incorporate a current and target network to stabilize learning. In DQN, the current network $\varpi^{Q}_{n_c}$, known as Q-network, is responsible for jointly selecting and evaluating discrete actions. In contrast, DDPG employs an actor-critic structure, where the current actor network $ \omega^\mu_{n_c} $ generates continuous actions and the current critic network $ \omega^Q_{n_c} $ evaluates the corresponding Q-values. Both the target Q-network of DQN $\varpi^{Q'}_{n_c}$ and target actor/critic networks of DDPG $ \{ \omega^{\mu'}_{n_c}, \omega^{Q'}_{n_c} \}$ serve to stabilize the learning process by periodically copying the models from their respective current networks. The actions are generated by exploration-exploitation policy in current networks to prevent local solutions: (1) Each DQN agent selects the best action by either its Q-function ${Q}_{n_c}^{\text{dis}}(\cdot)$ as $ a_{n_c}^{\text{dis}}(t) = \argmax_{a^{\text{dis}'}_{n_c}(t) \in \mathcal{A}^{\text{dis}}_{n_c} } {Q}_{n_c}^{\text{dis}} (s_{n_c}(t),a^{\text{dis}'}_{n_c}(t) | \varpi^Q_{n_c})$ if $\varsigma \geq \varsigma_{\text{th}}$ or random selection otherwise, where $\varsigma$ is a randomly generated value and $\varsigma_{\text{th}}$ is the exploration threshold; (2) Each DDPG agent determines its action by $ a_{n_c}^{\text{con}}(t) = \mu_{n_c}(s_{n_c}(t) | \omega^\mu_{n_c}) + \chi $, where the first term indicates the action output of the current actor network $\omega^\mu_{n_c}$ given the input state $s_{n_c}(t)$. Notation of $\chi$ is defined as Gaussian noise for exploring potential new actions in the environment. The reward $r_{n_c}(t)$ will be obtained and the state will be updated as $s_{n_c}(t+1) = s_{n_c}(t)$ after executing the current networks' actions. 

The memory replay buffer $\mathcal{D}$ with a size of $|\mathcal{D}|$ is constructed to store the trajectory, i.e., historical experiences, with the tuple of $\Xi_{n_c}^v(t) = (s_{n_c}(t), a^{v}_{n_c}(t), r_{n_c}(t), s_{n_c}(t+1))$ where $v\in\{\text{dis}, \text{con} \}$. We randomly select $I \leq |\mathcal{D}|$ samples from $\mathcal{D}$ for the neural network training, where the data sample set is defined as $\mathcal{I}=\{1,2,...,I\}$. Based on data at time instant $i \in  \mathcal{I}$, the corresponding target values of MADQN and MADDPG are acquired as
\begingroup
\allowdisplaybreaks
\begin{subequations}
\begin{align}
    y_{\text{tar}}^{\text{dis}} &= r_{n_c}(i) \!+\! \gamma \cdot \max_{a_{n_c}^{\text{dis}'}(i) \in\mathcal{A}^{\text{dis}}_{n_c} }{Q}_{n_c}^{\text{dis}'} \left(s_{n_c}(i \!+\! 1), a_{n_c}^{\text{dis}'}(i) | \varpi^{Q'}_{{n_c}} \right),  \label{value_dqn} \\
    y_{\text{tar}}^{\text{con}} &= r_{n_c}(i) + \gamma \cdot {Q}_{n_c}^{{\text{con}}'} \left(s_{n_c}(i+1), a^{\text{con}'}_{n_c}(i) | \omega^{Q'}_{n_c} \right), \label{value_ddpg}
\end{align}
\end{subequations}
\endgroup
where ${Q}_{n_c}^{\text{dis/con}'}(\cdot)$ indicates the Q-function of MADQN/MADDPG target networks. Notations of $a_{n_c}^{\text{dis}'}(i)$ and $a^{\text{con}'}_{n_c}(i)$ present the respective actions generated by the MADQN/MADDPG target networks, following the same generation process. Also, the discount factor denoted as $ \gamma \in [0, 1] $ indicates the importance of future rewards. To minimize the estimation error between target and the current networks, the general loss functions of MADQN and of MADDPG are formulated as
\begin{align}
    \mathcal{L}_{Q} =  \mathbbm{E}_{\Xi_{n_c}^v(i)}\left[ \left(y_{\text{tar}}^{v} - Q_{n_c}^{v} \left(s_{n_c}(i), a_{n_c}^{v}(i) | \tilde{\omega}^{Q}_{n_c} \right) \right)^2 \right],\label{dqnloss}
\end{align}
where $\tilde{\omega} \in \{ \varpi,  \omega \}$. Note that this loss in DDPG is also termed as critic loss. Moreover, DDPG conducts additional actor loss minimization for higher rewards as 
\begin{align} \label{ddpgaloss}
	\mathcal{L}_{\text{act}} = -\mathbbm{E}_{s_{n_c}(i)} \left[ Q_{n_c}^{\text{con}} \left(s_{n_c}(i), a_{n_c}^{\text{con}}(i) | \omega^{Q}_{n_c} \right) \right]. 
\end{align}
The stochastic gradient descent is employed to update the model weights of the current network, i.e., $\tilde{\omega}^{Q}_{n_c} \leftarrow \tilde{\omega}^{Q}_{n_c} - \eta_{1} \cdot \nabla_{\tilde{\omega}^{Q}_{n_c}} \mathcal{L}_{Q}$ and $\omega^{\mu}_{n_c} \leftarrow \omega^{\mu}_{n_c} - \eta_{2} \cdot \nabla_{\omega^{\mu}_{n_c}} \mathcal{L}_{\text{act}}$, where $\eta_{1}, \eta_{2}$ are learning rates. The gradient of Q-loss of MADDPG/MADQN in \eqref{dqnloss} is calculated as $\nabla_{\tilde{\omega}^{Q}_{n_c}} \mathcal{L}_{Q} \approx \mathbbm{E} [ 2 (y_{\text{tar}}^{v} - Q_{n_c}^{v} (s_{n_c}(i), a_{n_c}^{v}(i) | \tilde{\omega}^{Q}_{n_c}) ) \cdot \nabla_{\tilde{\omega}^{Q}_{n_c}} Q_{n_c}^{v} (s_{n_c}(i), a_{n_c}^{v}(i) | \tilde{\omega}^{Q}_{n_c}) ]$. Moreover, the associated gradient of MADDPG's actor network in \eqref{ddpgaloss} is derived by the chain rule as $\nabla_{\omega^{\mu}_{n_c}} \mathcal{L}_{\text{act}} \approx \mathbbm{E}_{s_{n_c}(i)} [ \nabla_{a^{\text{con}}_{n_c}(i)} Q_{n_c}^{\text{con}} (s_{n_c}(i), a^{\text{con}}_{n_c}(i) | \omega^{Q}_{n_c} ) \cdot \nabla_{\omega^{\mu}_{n_c}} \mu_{n_c} (s_{n_c}(i) \mid \omega^{\mu}_{n_c} ) ]$. To elaborate further, the target network of MADQN/MADDPG will periodically update the neural network model weights from the current network based on the soft update \cite{17} as
\begingroup
\allowdisplaybreaks
\begin{subequations} \label{softup}
\begin{align}
   \tilde{\omega}_{n_c}^{ Q'} &\leftarrow \tau_{Q} \tilde{\omega}^{Q}_{n_c} + (1 - \tau_{Q}) \tilde{\omega}_{n_c}^{ Q'} \\
   \omega_{n_c}^{\mu'} &\leftarrow \tau_{\mu} \omega^{\mu}_{n_c} + (1 - \tau_{\mu}) \omega_{n_c}^{\mu'},
\end{align}
\end{subequations}
\endgroup
where $\tilde{\omega} \in \{ \varpi,  \omega \}$. Notations $ 0\leq \tau_{\mu} $ , $ \tau_{Q}\leq 1 $ denote positive constants indicating the portion of actor/critic parameters contributed by the current networks, respectively.

\subsection{Bi-Directional Parametrized Sharing}

In conventional multi-agent DRL architecture, all agents typically rely on the only feedback of rewards from the environment to adjust their policies. In hybrid DRL, two distinct neural network models handle different types of actions. The lack of proper information exchange can lead to policy misalignment, ultimately degrading overall system performance. Inspired by works of \cite{pdqn} and \cite{han}, we design bi-directional parametrized sharing to enable MADQN and MADDPG to learn from each other's selected parameters during the decision-making process, as depicted in Fig. \ref{MAS-H}. Specifically, the action selected by MADQN module will be fed back to MADDPG as additional input information. This enables MADDPG to account for the impact of discrete actions when determining its continuous parts. Similarly, continuous actions decided by MADDPG will be transmitted as auxiliary information as input of DQN, allowing it to incorporate continuous control influence into its discrete decision-making process. With the parametrized sharing mechanism, the action of MADQN and MADDPG of node $n_c$ at time $t$  is designed as
\begingroup
\allowdisplaybreaks
\begin{subequations}
\begin{align}
	a_{n_c}^{\text{dis}}(t) &= \argmax_{a^{\text{dis}'}_{n_c}(t) \in \mathcal{A}^{\text{dis}}_{n_c} } 
Q^{\text{dis}}_{n_c} \left(O^{\text{dis}}_{n_c}(t),a^{\text{dis}'}_{n_c}(t) | \varpi^Q_{n_c} \right), \\ 
   a_{n_c}^{\text{con}}(t)& = \mu_{n_c} \left( O^{\text{con}}_{n_c}(t) | \omega^\mu_{n_c} \right) + \chi,
\end{align}
\end{subequations}
\endgroup
where the input of MADQN/MADDPG is designed as the concatenation of the observed state and the previous action of MADDPG/MADQN given by $ O^{\text{dis}}_{n_c}(t) = {\rm Concat}(s_{n_c}(t), a^{\text{con}}_{n_c}(t-1))$ and $O^{\text{con}}_{n_c}(t) = {\rm Concat}(s_{n_c}(t), a^{\text{dis}}_{n_c}(t-1))$, respectively. Note that ${\rm Concat}(\cdot)$ represents the concatenation operation.

\subsection{VAE-based Semantic Compression of State-Action} 

\begin{figure}[!t]
\centering
\includegraphics[width=3.3in]{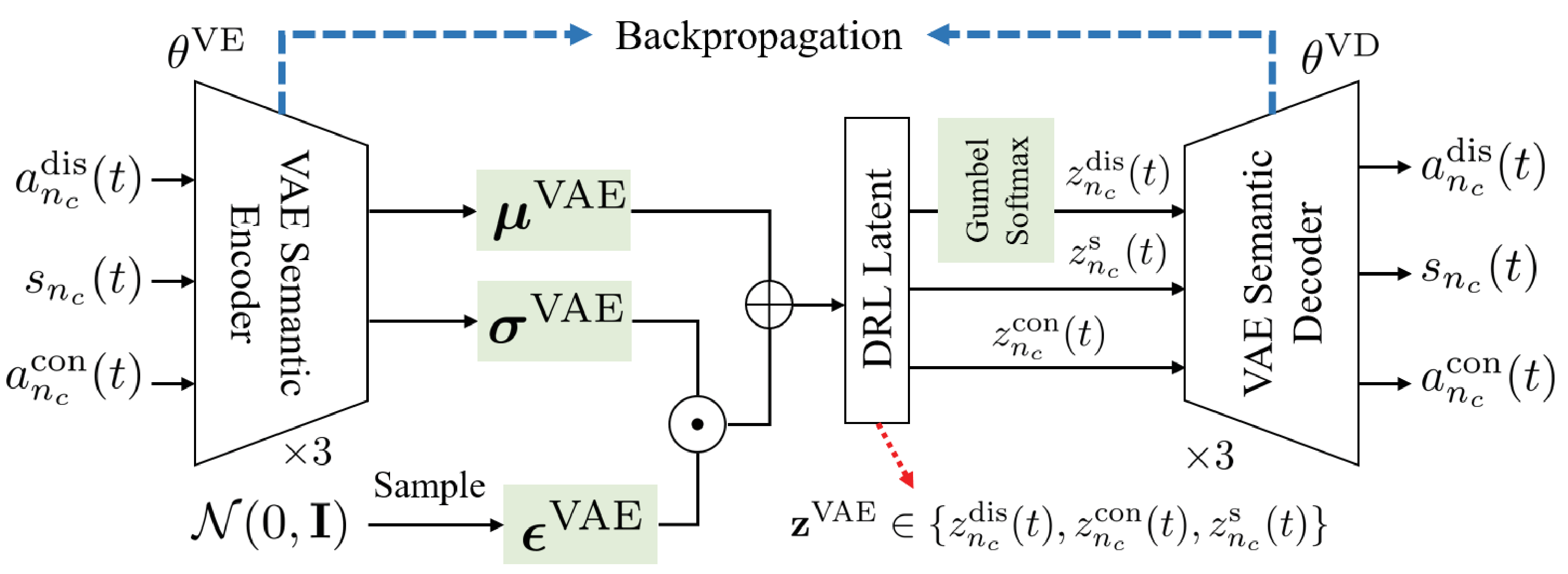}
\caption{VAE-based semantic compression for states and discrete/continuous actions. Note that three independent VAE is established for respective utilizations.} 
\label{compress}
\end{figure}

In the proposed SAGIN-MF-RIS network, there exist a large number of parameters posing a significant challenge for convergence and efficiency of DRL training. To circumvent these issues, inspired by \cite{my5}, we incorporate VAE-based semantic compression\textsuperscript{\ref{note3}}\footnotetext[3]{When dynamic environments are encountered, including unseen channels or network topologies. By employing knowledge distillation and transfer learning \cite{new_tf}, the pre-trained VAE models can be fine-tuned and generalized to new scenarios. Also, it can prevent massive data collections and VAE training from scratch, with reduced time overhead.\label{note3}} to extract meaningful states and actions but with substantially lower dimension.

\subsubsection{Continuous Action and State Compression}

The VAE uses the reparameterization trick with the latent defined as
\begin{align} \label{vaee}
    \mathbf{z}^{\text{VAE}} = \boldsymbol{\mu}^{\text{VAE}}(\mathbf{x}) + \boldsymbol{\sigma}^{\text{VAE}}(\mathbf{x}) \odot \boldsymbol{\epsilon}^{\text{VAE}},
\end{align}
where $\boldsymbol{\epsilon}^{\text{VAE}} \sim \mathcal{N}(0,\mathbf{I})$. Here, $\boldsymbol{\mu}^{\text{VAE}}(\mathbf{x})$ and $\boldsymbol{\sigma}^{\text{VAE}}(\mathbf{x})$ indicate the mean and standard deviation of the latent space $\mathbf{z}^{\text{VAE}}\in\{ z^{\text{s}}_{n_c}(t), z^{\text{con}}_{n_c}(t), z^{\text{dis}}_{n_c}(t) \}$ learned from input $\mathbf{x}\in\{ s_{n_c}(t), a^{\text{con}}_{n_c}(t), a^{\text{dis}}_{n_c}(t) \}$, respectively. Note that only the continuous actions and states can be directly encoded as \eqref{vaee} due to its differentiability. The associated encoded processes respectively given by $z^{\text{s}}_{n_c}(t) = {\rm VE}(s_{n_c}(t); \theta^{\rm VE})$ and $z^{\text{con}}_{n_c}(t) = {\rm VE}(a^{\text{con}}_{n_c}(t); \theta^{\rm VE})$, where ${\rm VE}(\cdot)$ is the encoder function and $\theta^{\rm VE}$ is the corresponding neural network. In contrast, the VAE decoder model $\theta^{\rm VD}$ is responsible for recovering the original state and actions as $s_{n_c}(t) = {\rm VD}(z^{\text{s}}_{n_c}(t); \theta^{\rm VD})$ and $a^{\text{con}}_{n_c}(t) = {\rm VD}(z^{\text{con}}_{n_c}(t); \theta^{\rm VD})$, where ${\rm VD}(\cdot)$ is the decoder function.

\subsubsection{Discrete Action Compression}

Since discrete variables are non-differentiable during backpropagation training, we employ the Gumbel-softmax VAE \cite{gsv} for approximating the compressed discrete actions to categorical sampling, given by
\begin{align} \label{vae_dis}
    z^{\text{dis}}_{n_c}(t) \!=\! {\rm GVE}(a^{\text{dis}}_{n_c}(t)) = \frac{\exp \left( (\log \pi_i + g_i) / \tau^{\text{VAE}} \right)}{\sum_{j} \exp\left( (\log \pi_j + g_j) / \tau^{\text{VAE}} \right)},
\end{align}
where ${\rm GVE}(\cdot)$ is the Gumbel-softmax encoder function and $\pi_i$ represents the categorical probabilities. Notation $g_i \sim {\rm Gumbel}(0,1)$ presents a random perturbation from the standard Gumbel distribution with location 0 and scale 1. Notation $\tau^{\text{VAE}}$ is the temperature parameter controlling the continuous property, i.e., the output of the softmax function with $\tau^{\text{VAE}} \to 0$ approaches a one-hot vector. This ensures that discrete actions remain distinct while still allowing gradient-based optimization. Moreover, the recovery of the discrete action follows the same process as that of continuous case, which is $a^{\text{dis}}_{n_c}(t) = {\rm VD}(z^{\text{dis}}_{n_c}(t); \theta^{\rm VD})$.

The loss function of VAE consists of a reconstruction loss and a Kullback-Leibler (KL) divergence penalty $D_{\text{KL}}(\cdot)$ as
\begin{align} \label{vaeloss}
    \mathcal{L}_{\text{VAE}} &= \mathbb{E}_{q(\mathbf{z}^{\text{VAE}}|\mathbf{x}^{\text{VAE}})}  \left[\log p(\mathbf{x}^{\text{VAE}}| \mathbf{z}^{\text{VAE}}) \right] \notag \\ & \qquad\qquad
    - D_{\text{KL}}\left( q( \mathbf{z}^{\text{VAE}}|\mathbf{x}^{\text{VAE}}) \parallel p(\mathbf{z}^{\text{VAE}}) \right),
\end{align}
where $p(\mathbf{x}^{\text{VAE}}|\mathbf{z}^{\text{VAE}})$ indicates the likelihood corresponding to the reconstructed data given the latent representation. While, $q(\mathbf{z}^{\text{VAE}}|\mathbf{x}^{\text{VAE}})$ denotes the approximate Gaussian posterior. Note that $p(\mathbf{z}^{\text{VAE}}) \sim \mathcal{N}(0, \mathbf{I})$ is the prior distribution of the latent variable, typically following the normal distribution. To avoid cross-contamination between state and action representations, we employ three separate VAE modules trained independently. To elaborate further, the training process is conducted offline, ensuring that the VAE model is well-trained prior to executing the hybrid DRL framework. During the online phase of CHIMERA, the encoder and decoder of the VAE are respectively positioned before the input and after the output of hybrid DRL, as illustrated in Fig. \ref{TMA}. Such method can accelerate the major learning process of DRL.

\subsection{CHIMERA with Twin-Models}
\begin{figure}[!t]
\centering
\includegraphics[width=3.3in]{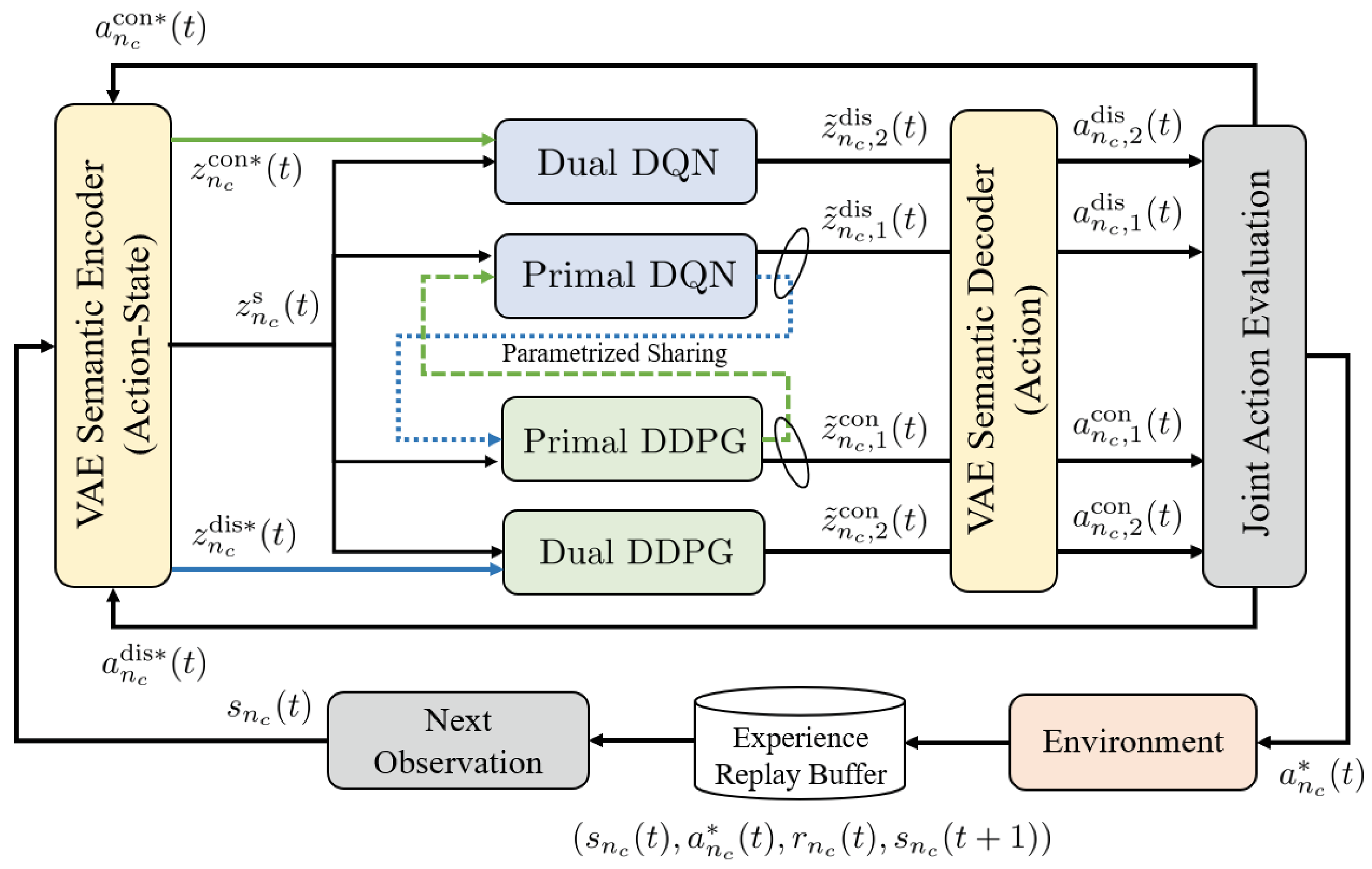}
\caption{The proposed CHIMERA framework with twin-models and VAE-based semantic state-action compression mechanisms.} 
\label{TMA}
\end{figure}

As shown in Fig. \ref{TMA}, inspired by double/dueling DQN \cite{dqn1, dqn2} utilizing two modules to prevent policy overfitting, the proposed CHIMERA framework is designed by incorporate twin-models into VAE-enabled hybrid DRL. It constitutes two pairs of hybrid DRL, including primal and dual DQN/DDPG neural network models of each agent. Following similar design of parametrized sharing, the dual DQN/DDPG models' inputs include information of both the commonly observed state $s_{n_c}(t)$ and the individual best action $ a_{n_c}^{\text{con}*}(t) / a_{n_c}^{\text{dis}*}(t)$ of DDPG/DQN previously evaluated. Here, after VAE semantic encoder, the four respective inputs of primal/dual DQN/DDPG become 
	$O^{\text{dis}}_{n_c,1}(t) = {\rm Concat}(z^{\text{s}}_{n_c}(t), \tilde{z}_{n_c,1}^{\text{con}}(t))$,
	$O^{\text{dis}}_{n_c,2}(t) = {\rm Concat}(z^{\text{s}}_{n_c}(t), z^{\text{con}*}_{n_c}(t))$,
	$O^{\text{con}}_{n_c,1}(t) = {\rm Concat}(z^{\text{s}}_{n_c}(t), \tilde{z}^{\text{dis}}_{n_c,2}(t))$, and
	$O^{\text{con}}_{n_c,2}(t) = {\rm Concat}(z^{\text{s}}_{n_c}(t), \tilde{z}^{\text{dis}*}_{n_c}(t))$, where the respective compressed actions of primal/dual DQN/DDPG are designed as
\begingroup
\allowdisplaybreaks
\begin{subequations}
\begin{align}
	\tilde{z}_{n_c,i}^{\text{dis}}(t) & \!=\! \argmax_{ z_{n_c,i}^{\text{dis}'}(t) \in \mathcal{Z}_{n_c}^{\text{dis}}} Q_{n_c,i}^{\text{dis}} \big( O^{\text{dis}}_{n_c,i}(t), z_{n_c,i}^{\text{dis}'}(t) | \varpi_{n_c,i}^{Q} \big), \\
	\tilde{z}_{n_c,i}^{\text{con}}(t) &= \mu_{n_c,i} \big(O^{\text{con}}_{n_c,i}(t) | \omega_{n_c,i}^{\mu} \big) + \chi, \  \forall i\in\{1,2\},
\end{align}
\end{subequations}
\endgroup
where $\mathcal{Z}_{n_c}^{\text{dis}} = {\rm VE}(\mathcal{A}_{n_c}^{\text{dis}}; \theta^{\rm VE})$ is the compressed total discrete action set. The compressed actions will then be recovered to its original action through VAE semantic decoder, i.e., ${a}_{n_c,i}^v(t) = {\rm VD}(\tilde{z}_{n_c,i}^v(t);\theta^{\rm VD}), \forall v\in\{\text{dis}, \text{con}\}, \forall i\in\{1,2\}$. Note that subscripts $i=1$ and $i=2$ represent the primal and dual networks, respectively. Afterwards, the intermediate decisions made by DQN and DDPG after VAE decoder will be combined to form a final best action strategy, denoted as $ a_{n_c}^{*}(t)= \{ a_{n_c}^{\text{con}*}(t), a_{n_c}^{\text{dis}*}(t) \} = \argmax_{ (a^{\text{con}}_{n_c} (t), a^{\text{dis}}_{n_c} (t)) \in \Xi_{\text{act}}} r_{n_c}(t) $. Note that we adopt a combinatorial optimization approach to exhaustively obtain the best action associated with the highest reward, i.e., total four possible combinations of $\Xi_{\text{act}} = \{ (a^{\text{con}}_{n_c,i_1}(t), a^{\text{dis}}_{n_c,i_2}(t) | \forall (i_1,i_2) \in\{ (1, 1), (1, 2), (2, 1), (2, 2) \} \}$. It is worth noting that the parametrized sharing mechanism improves training diversity, allowing the primal-dual architecture to explore more diverse solution spaces.

The overall procedure of CHIMERA is elaborated in Algorithm \ref{alg}. The process begins with initializing the SAGIN environment and CHIMERA network modules, which is followed by the preparation of pre-trained VAE. Each SAGIN node is then assigned empty action sets, replay buffers, and mini-batch sets. At each time step, every SAGIN node selects actions from its CHIMERA actor network, where the states and actions can optionally be compressed using the pre-trained VAE module. After executing the selected actions, all nodes observe their updated states and obtain the corresponding shared reward. Subsequently, MADQN/MADDPG-based training and model weight updates are performed. The iterative process above continues until the traversal of the entire SAGIN system is completed.

\begin{algorithm}[!t]
\caption{Proposed CHIMERA Scheme}
\label{alg}
\begin{algorithmic}[1]
\STATE \textbf{Initialization:}
\STATE \quad 1) Initialize environment, DQN/DDPG networks
\STATE \quad 2) Set $a^{\text{dis}}_{n_c}, a^{\text{con}}_{n_c}, a^{\text{best}}_{n_c} \leftarrow \mathbf{0}$
\STATE \quad 3) Pre-train VAE and Gumbel-Softmax VAE based on \eqref{vaee}--\eqref{vaeloss}
\STATE \quad 4) Initialize replay buffer $\mathcal{D}$ and mini-batch set $\mathcal{X}$

    \FOR{$t = 1$ to $T$}
        \FOR{$i = 1$ to $N_C$}
        	\STATE Apply VAE in \eqref{vaee} and \eqref{vae_dis} for states if compression is required
            \STATE $a^{\text{dis}}_{n_c,1} \gets$ Primal-DQN$({\rm Concat}(s_{n_c}(t) , a^{\text{con}}_{n_c}(t) ))$
            \STATE $a^{\text{dis}}_{n_c,2} \gets$ Dual-DQN$({\rm Concat}(s_{n_c}(t) , a^{\text{con}*}_{n_c}(t)))$
            \STATE $a^{\text{con}}_{n_c,1} \gets$ Primal-DDPG$({\rm Concat}(s_{n_c}(t) , a^{\text{dis}}_{n_c} (t)))$
            \STATE $a^{\text{con}}_{n_c,2} \gets$ Dual-DDPG$({\rm Concat}(s_{n_c}(t), a^{\text{dis}*}_{n_c}(t)))$
            \STATE Apply VAE in \eqref{vaee} and \eqref{vae_dis} for actions above if compression is required
            \STATE Apply joint action $a_{n_c}^{*}(t)$ to environment
        \ENDFOR

        \FOR{$n_c = 1$ to $N_c, \forall c\in\mathcal{C}$ }
            \STATE Observe $s_{n_c}(t+1)$ and acquire reward $r_{n_c}(t)$
            \STATE Store $\Xi_{n_c}^v(t)$ in $\mathcal{D}$
            \STATE Sample mini-batch $\mathcal{I}$ from $\mathcal{D}$

            \STATE \textbf{MADQN Module:}
            \STATE Compute target value in \eqref{value_dqn}
            \STATE Compute Q loss based on \eqref{ddpgaloss}
			\STATE Perform gradient descent to update MADQN parameters
			\STATE Soft update based on \eqref{softup}
            \STATE \textbf{MADDPG Module:}
            \STATE Compute target value in \eqref{value_ddpg}
            \STATE Compute actor/critic loss in \eqref{ddpgaloss}/\eqref{dqnloss} to obtain gradients and update parameters
            \STATE Soft update based on \eqref{softup}
        \ENDFOR
    \ENDFOR
\end{algorithmic}
\end{algorithm}

\section{Simulation Results} \label{sec_sim}

\begin{table}[!t]
\centering
\footnotesize
\caption{Simulation Parameters}
\label{tab:sim_params}
\renewcommand{\arraystretch}{1.2}
\begin{tabular}{|l|l|}
\hline
\textbf{Channel Parameters} & \textbf{Values} \\
\hline
Number of Nodes $N_c$ & 6 \\
Number of IoT devices $K$ & 18 \\
Number of Antennas $N$ & 16 \\
Number of MF-RIS Elements $M$ & 32 \\
Transmit Power $P^{\text{max}}_{n_c}$ (LEO/HAPS/BS) & 60, 40, 30 dBm \\
Pathloss Reference Gain $h_0$ & $-20$ dB \\
Pathloss Exponent $k_0$ & 2.2 \\
Rician Factor $\beta_0$ & 5 \\
AWGN Power $\sigma_{n_c,k}^2$, $\sigma_m^2$ & $-80$ dBm \\
\hline
\textbf{MF-RIS Parameters} & \\
\hline
Amplifier Efficiency $\xi$ & 1.1 \\
Maximum amplitude $\beta_{\text{max}}$ & [0.5, 6] \\
PIN Diode Power $P_{\text{pin}}$ & 0.33 mW \\
Conversion Circuit Power $P_C$ & 10 W \\
Control Power $P_{n_c}^{\text{cons}}$ & 90 W \\
EH Models $\{Z,a,q\}$ & 24 mW, 150, 0.014 \\
Sensitivity threshold $P^{\text{RF}}_{\text{th}}$ & 2 mW \\
\hline
\textbf{SAGIN Parameters} & \\
\hline
Quantization Levels $(L_\alpha, L_\beta, L_\theta)$ & ($2^2$, $2^{10}$, $2^8$) \\
Battery Capacity $E^b$ & $9 \times 10^4$ J \\
Solar Panel Efficiency $\eta_s$ & 0.19 \\
Light Intensity $\psi$ & 500 W/m$^2$ \\
Solar Panel Size $B$ & 4 m$^2$ \\
Earth Radius $R_e$ & 6378 km \\
LEO Altitude $h_{n_s}$ & 600 km \\
HAPS Altitude $h_{n_a}$ & 50 km \\
Earth Rotation Rate $\dot{\Omega}$ & $7.29 \times 10^{-5}$ rad/s \\
Computation Coefficient $\tau$ & $10^{-28}$\\
Data Size $D^{\text{comp}}_{n_c,k}$ & $64$ KB\\
\hline
\end{tabular}
\end{table}

In simulations, we construct the Globalstar LEO satellite system \cite{29} to evaluate the proposed SAGIN-MF-RIS performance. The pertinent parameters are listed in Table~\ref{tab:sim_params}. Note that the detailed terrestrial and non-terrestrial channel parameters and settings can be found in \cite{sim_ch1} and \cite{sim_ch2,sim_ch3}, respectively. As for the settings of the proposed CHIMERA scheme, MADDPG adopts learning rates of $\{0.0001, 0.0002\}$ for the actor and critic networks, respectively. The discount factor is set to $\gamma = 0.99$, and the soft update coefficients for both actor and critic target networks are $\tau_{\mu} = \tau_{Q} = 0.0001$. In MADQN, the learning rate is set to $0.001$, whereas the target network is softly updated with $\tau_{Q} = 0.01$. The exploration of MADQN is governed by a greedy policy, where exploration rate $\varsigma$ linearly decays from $1$ to $0.1$ over $10^4$ steps. The experience replay buffer $\mathcal{D}$ of MADQN/MADDPG can store up to $10^5$ samples, whilst both models are trained with a mini-batch size of $64$. Note that all neural networks are initially randomly generated to ensure sufficient diversity between the twin-models from the start of training. Note that using the same initial neural network model in both primal and dual networks is acceptable, as their input states are distinct.

\begin{table*}[!ht]
\centering
\footnotesize
\caption{Channel Loss of Multi-Reflections}
\label{pp1}
\renewcommand{\arraystretch}{1.1}
\begin{tabular}{|l|c|c|c|c|c|c|c|c|}
\hline
Channel Path & Direct & S & A & G & S-A & S-G & A-G & S-A-G \\
\hline
Averaged Loss (dB) & $110$ & $120$ & $172$ & $140$ & $182$ & $150$ & $202$ & $200$ \\
\hline
\end{tabular}
\end{table*}

\begin{table*}[!ht]
\centering
\footnotesize
\caption{Channel Loss of Reflection to Upper Layers}
\label{pp2}
\renewcommand{\arraystretch}{1.1}
\begin{tabular}{|l|c|c|c|c|c|c|c|c|c|c|c|c|}
\hline
Channel Path & \textbf{S-A-G} & S-A-S & S-G-S & S-G-A & A-S-A & A-S-G & A-G-S & A-G-A & G-S-A & G-S-G & G-A-S & G-A-G \\
\hline
Averaged Loss (dB) & $\mathbf{200}$ & $275$ & $278$ & $247$ & $327$ & $295$ & $330$ & $299$ & $330$ & $298$ & $330$ & $267$ \\
\hline
\end{tabular}
\end{table*}

In Table \ref{pp1}, we further evaluate that the intermediate simulations yield average channel losses of approximately $\{110, 144, 178, 200\}$ dB for the single- (i.e., direct link), double-, triple- and quadruple-bounce paths in the LEO scenario in \eqref{chh}. Nevertheless, such severe attenuation can be compensated by deploying a larger number of MF-RISs and antennas with higher amplification gains. In particular, the quintuple-bounce can effectively leverage three amplification stages enabled by the cascaded MF-RISs. Additionally, as listed in Table \ref{pp2}, we have quantitatively justified the difference between the case with reflections to upper layers and that without it in \eqref{chh}, i.e., LEO-S-A-G-Device where S/A/G indicates where MF-RIS is reflected. Considering the LEO channel, its paths involving reflections to upper layers include combinations of LEO $\rightarrow$ S-A-S, S-G-S, S-G-A, A-S-A, A-S-G, A-G-S, A-G-A, G-S-A, G-S-G, G-A-S, G-A-G $\rightarrow$ the IoT device. The LEO-S-A-G-Device case without upper-layer reflections exhibits a channel loss of approximately $200$ dB, whereas that including reflections to upper layers results in losses around $298$ dB. Hence, the channel paths involving upper-layer reflections is ignored due to their negligible contribution to the overall received power.

\subsection{Convergence of CHIMERA}

\begin{figure}[!t]
\centering
\includegraphics[width=3.3in]{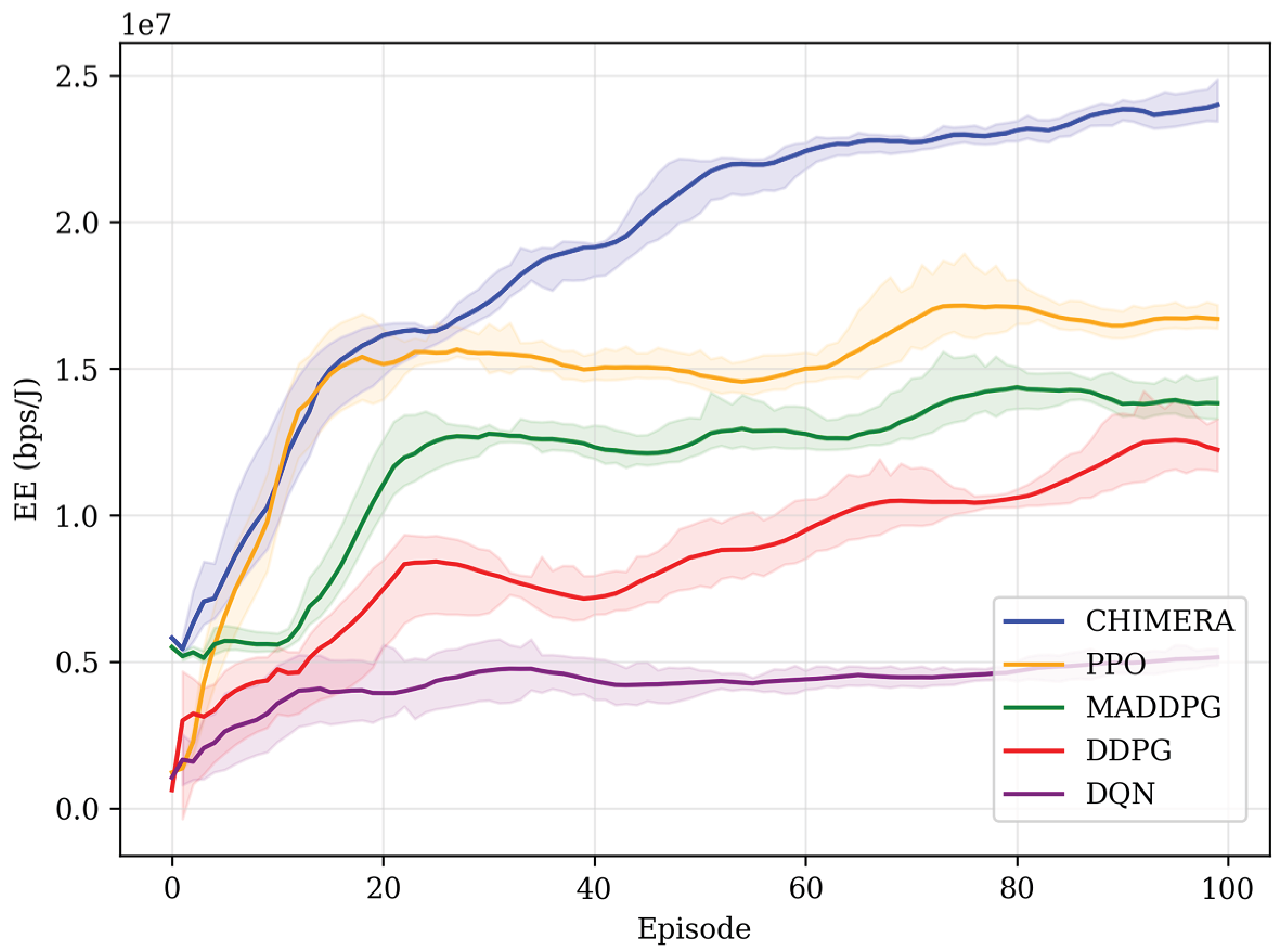}
\caption{Convergence of CHIMERA and existing DRL methods of DQN, DDPG, PPO, and MADDPG.} \label{f.sim.1}
\end{figure}

Fig. \ref{f.sim.1} illustrates the convergence behavior of the proposed CHIMERA framework against existing DRL benchmarks including proximal policy optimization (PPO), MADDPG, centralized learning of DDPG and DQN. Among the benchmarks, MADDPG benefited by inter-node coordination outperforms both DQN and DDPG as single-agent paradigm potentially limits its ability to capture inter-agent dependencies and often leads to poor policy generalization. Meanwhile, DQN is hindered by the curse of dimensionality due to its discrete states and actions, resulting in the lowest EE. Moreover, PPO is theoretically well-suited for handling mixed continuous-discrete action spaces; however, its exploration efficiency under centralized training deteriorates in high-dimensional environments. This limitation often necessitates larger neural network architectures to capture latent features, yet such complexity can cause severe overfitting and result in suboptimal policies due to single model and agent training. In contrast, the proposed CHIMERA framework effectively overcomes these limitations by integrating twin-model-based decision making, hybrid DRL structures, and compression techniques. As a result, CHIMERA achieves superior EE performance in large-scale and multi-agent environments with hybrid and high-dimensional state-action spaces.

\begin{figure}[!t]
\centering
\includegraphics[width=3.3in]{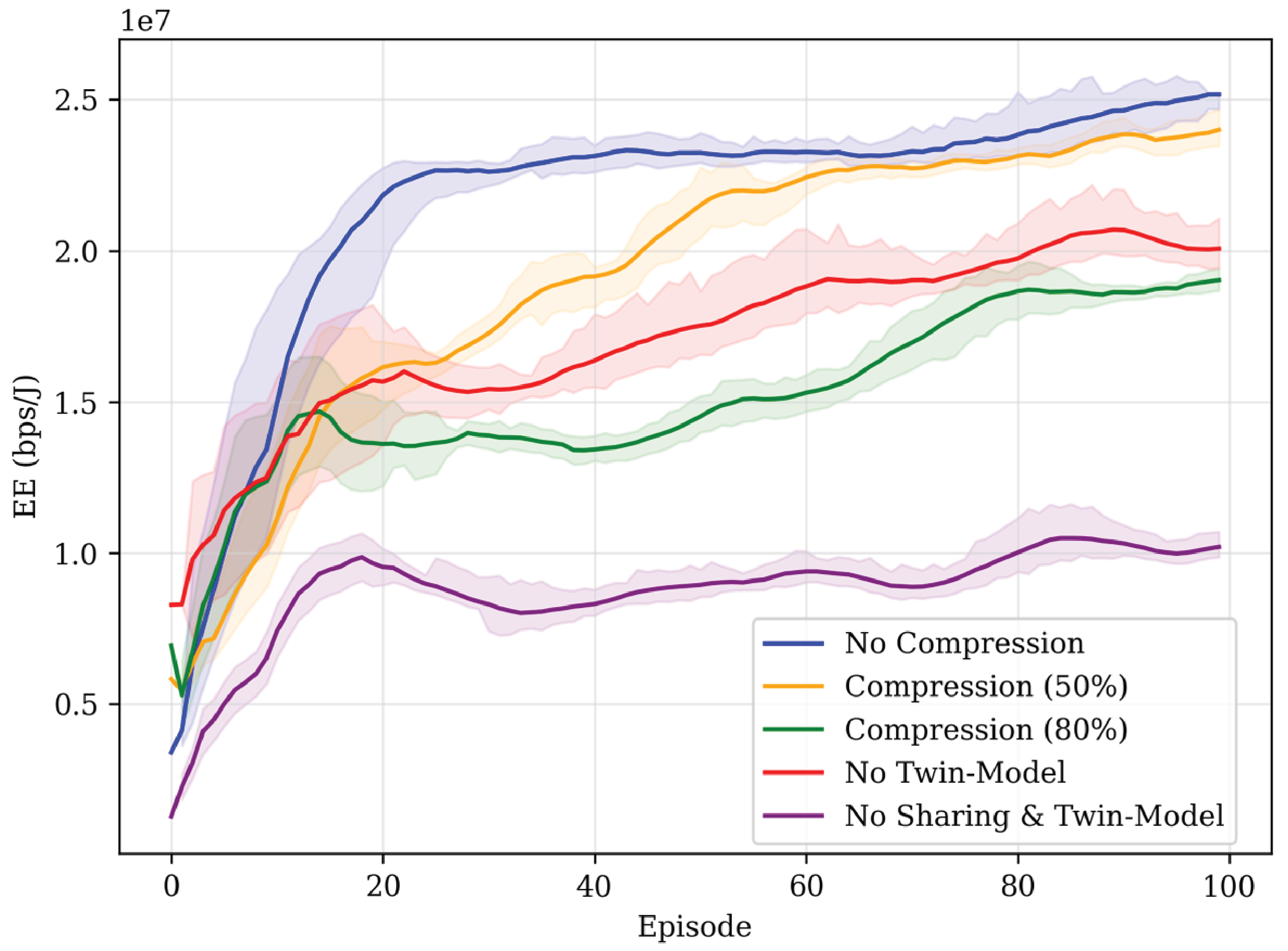}
\caption{Convergence of CHIMERA under different compression ratios and its reduced version of architectures.} \label{f.sim.2}
\end{figure}

\begin{figure}[!t]
\centering
\includegraphics[width=3.3in]{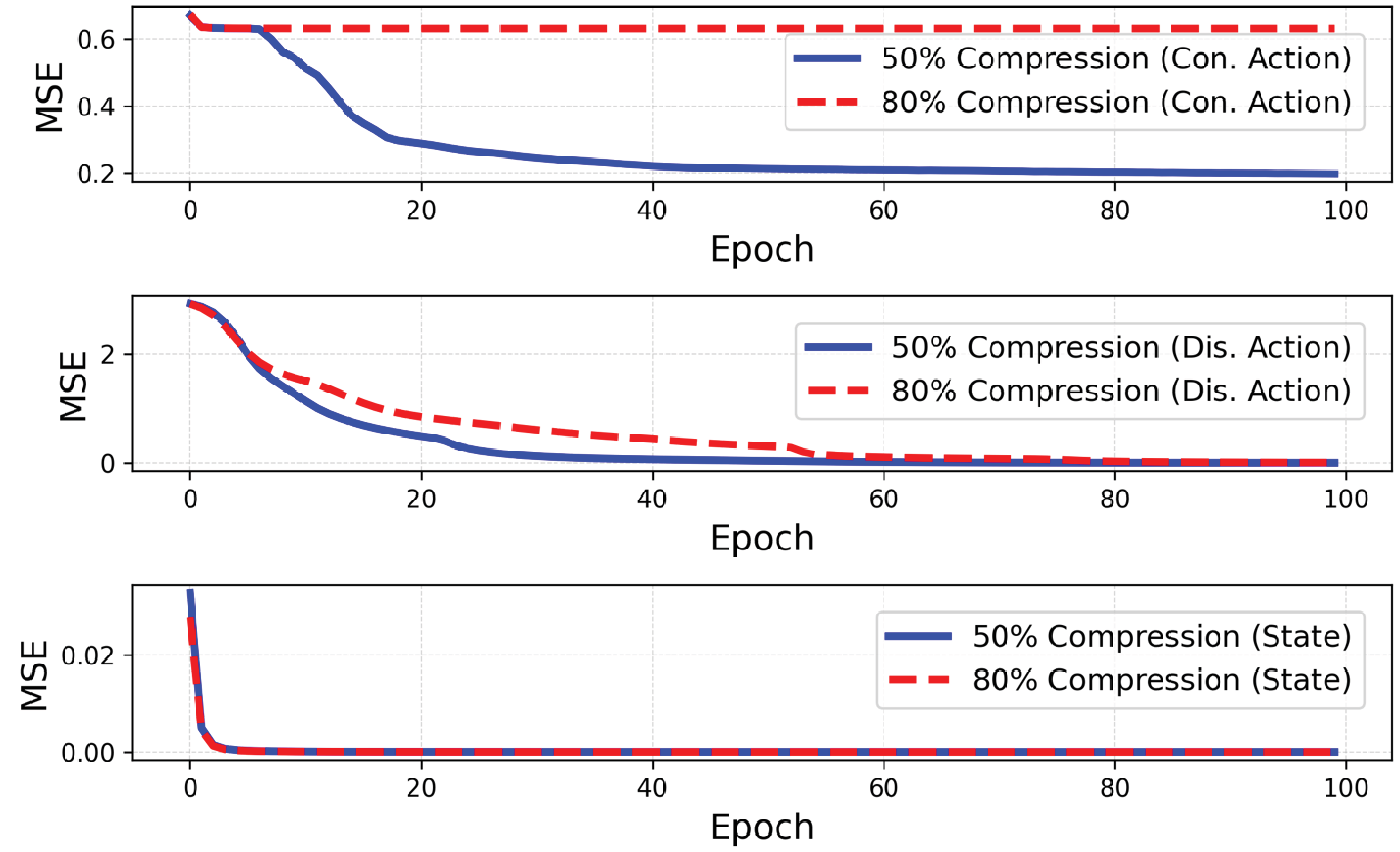}
\caption{Reconstruction MSE of the compressed continuous/discrete actions and states.} \label{mse_plot}
\end{figure}

Fig. \ref{f.sim.2} illustrates the convergence behavior of the proposed CHIMERA framework under varying compression ratios and highlights the impact of information sharing. As expected, the uncompressed CHIMERA attains the highest EE by retaining full training features. In comparison, a compression ratio of $50\%$ results in slightly slower convergence during the initial training phase but ultimately has just around $5\%$ EE degradation compared to the uncompressed case, while significantly reducing training overhead. However, excessive compression at $80\%$ incurs substantial information loss, leading to a noticeably declined EE of around $30\%$ compared to the case of $50\%$ compression. Furthermore, involving information sharing between the MADDPG and MADQN models yields a doubled EE, highlighting the benefits of policy exchange. Additionally, the twin-model architecture contributes to an approximate $25\%$ EE gain, further enhancing overall system performance through improved policy diversity. The corresponding mean square error (MSE) values of compression ratios $\{50\%, 80\%\}$ are depicted in Fig. \ref{mse_plot}. It can be observed that a higher compression ratio leads to a larger MSE, implying potential inaccurate DRL training in CHIMERA. Moreover, MSE difference between two compression ratios in the state representations is negligible, indicating that the compressed states still preserve sufficient latent features for accurate reconstruction.

\subsection{Effect of MF-RIS}


\begin{figure}[!t]
\centering
\includegraphics[width=3.3in]{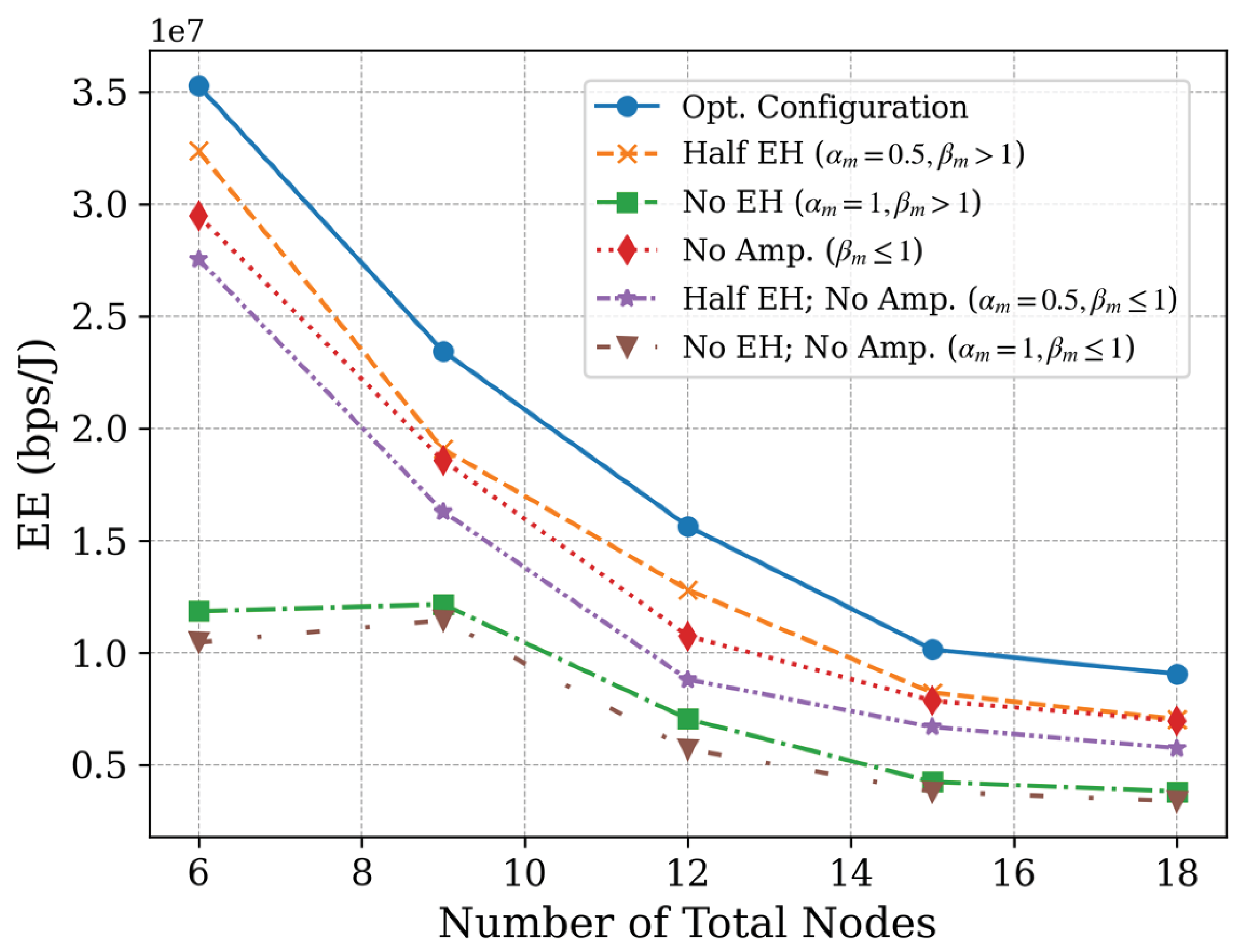}
\caption{EE performance with various MF-RIS configurations under different numbers of total nodes.} \label{f.sim.3}
\end{figure}

Fig. \ref{f.sim.3} compares the EE performance under different MF-RIS configurations, including (1) MF-RIS with optimized settings, (2) MF-RIS with half EH capability, (3) MF-RIS without EH functionality, (4) MF-RIS without signal amplification, and (5) conventional RIS features without both EH and signal amplification. The x-axis denotes the total number of SAGIN nodes, with an equal number of nodes allocated across each network layer. Note that we omit subscript $n_c$ here for simplicity. The results indicate that when EH is disabled $\alpha_{m}=1$, EE initially exhibits a slight improvement with the increasing number of nodes, but subsequently declines significantly. This trend suggests that excessive node deployment leads to substantially higher energy consumption, potentially degrading system EE. In contrast, enabling EH $\alpha_{m}\neq 1$ yields significantly higher EE compared to non-EH cases, implying that the self-sustaining capability of MF-RIS effectively supports both signal amplification and energy harvesting functions, even under limited deployment. However, as the number of nodes grows, the total power demand may exceed the harvested energy, resulting in a drop in EE performance. For the case of half EH $\alpha_m = 0.5$, the system achieves an improved EE by leveraging partially harvested energy to support enhanced signal amplification $\beta_m>1$, outperforming the case without amplification $\beta_m\leq 1$. Additionally, optimizing the EH ratio $\alpha_m$ can provide a higher EE compared to fixed configurations. In summary, jointly optimizing signal amplification factors and EH coefficients can significantly enhance system rate while minimizing energy consumption, emphasizing the importance of a balanced and adaptive MF-RIS configuration strategy.

\begin{figure}[!t]
\centering
\includegraphics[width=3.3in]{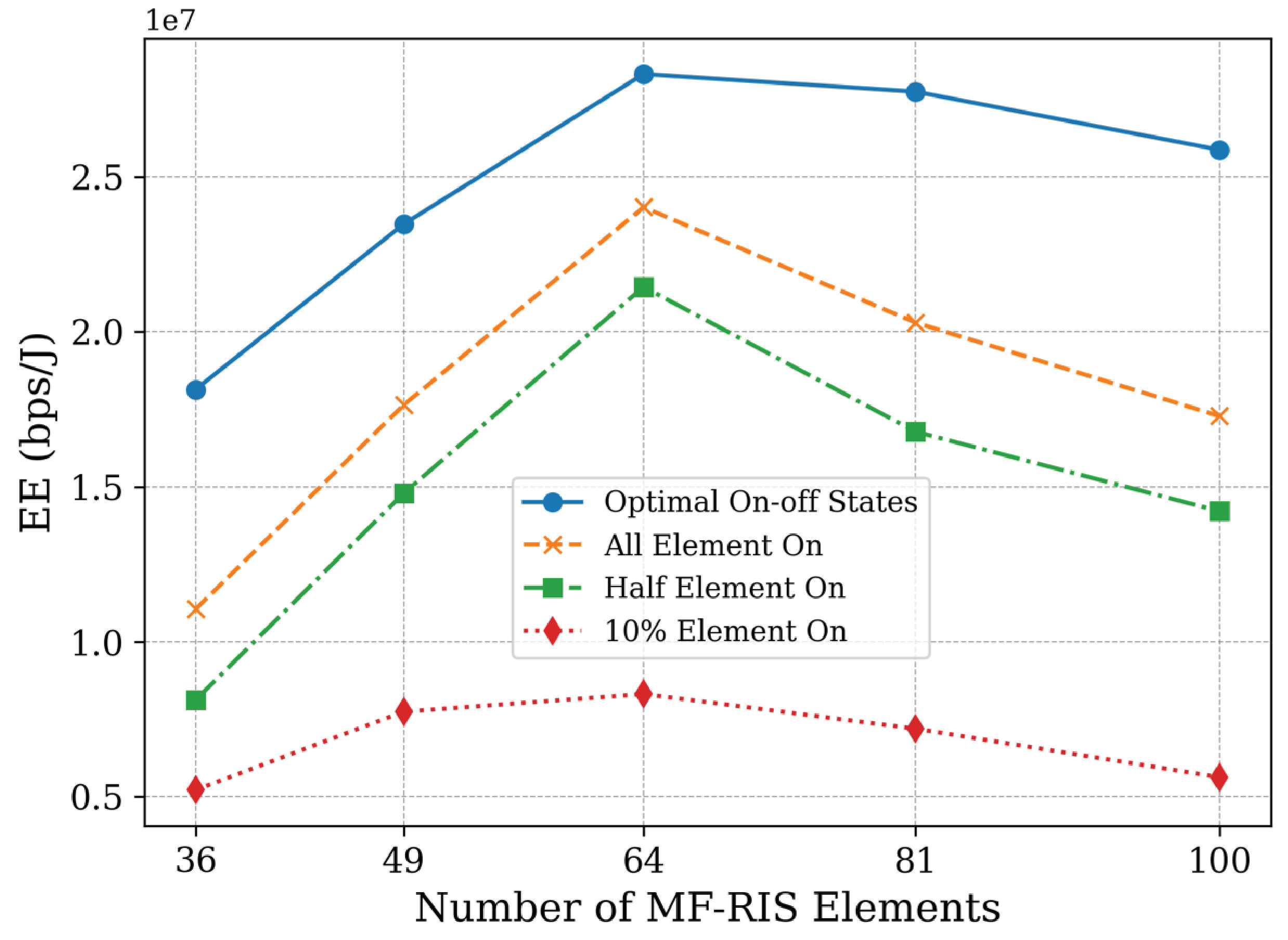}
\caption{EE performance with partial elements selected to be on under operation under different numbers of MF-RIS elements} \label{f.sim.4}
\end{figure}

Fig. \ref{f.sim.4} reveals EE performance under varying numbers of MF-RIS elements. Initially, EE increases with the number of MF-RIS elements, as additional elements enhance signal amplification flexibility and EH capabilities. The performance peaks at $M=64$, where the balance between signal gain and energy consumption is optimized. Beyond this point, EE declines as the power overhead associated with activating more elements outweighs the marginal benefits from EH. Furthermore, we compare the optimized case with baseline scenarios where a fixed portion of elements are randomly activated, including 10\%, half-on, and full-on configurations. The results reveal that the optimal on-off case selectively deactivating elements under suboptimal conditions can improve data rate and reduce unnecessary power consumption, thereby achieving the highest EE. It is worth noting that the full-on case outperforms the half-on configuration due to increased flexibility in beam control and EH. In contrast, the lowest EE is observed in the case of only 10\% of elements selected, where limited amplification and EH capacity significantly constrain system performance.

\begin{figure}[!t]
\centering
\includegraphics[width=3.3in]{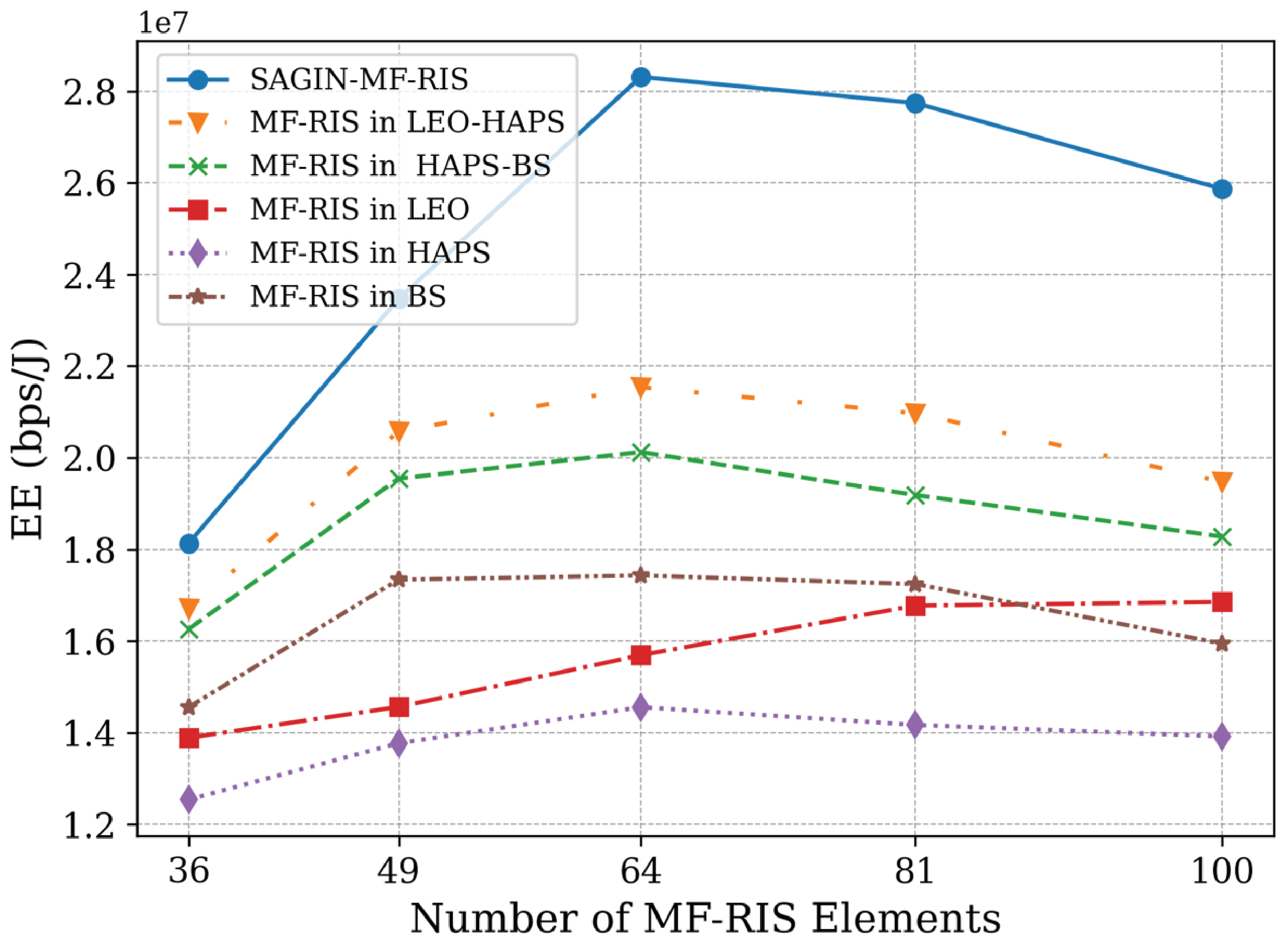}
\caption{EE performance versus different numbers of MF-RIS elements under various SAGIN layers deployed with MF-RIS.} \label{f.sim.7}
\end{figure}

\begin{figure*}[!t]
\centering
\subfigure[]
{\includegraphics[width=2.2in]{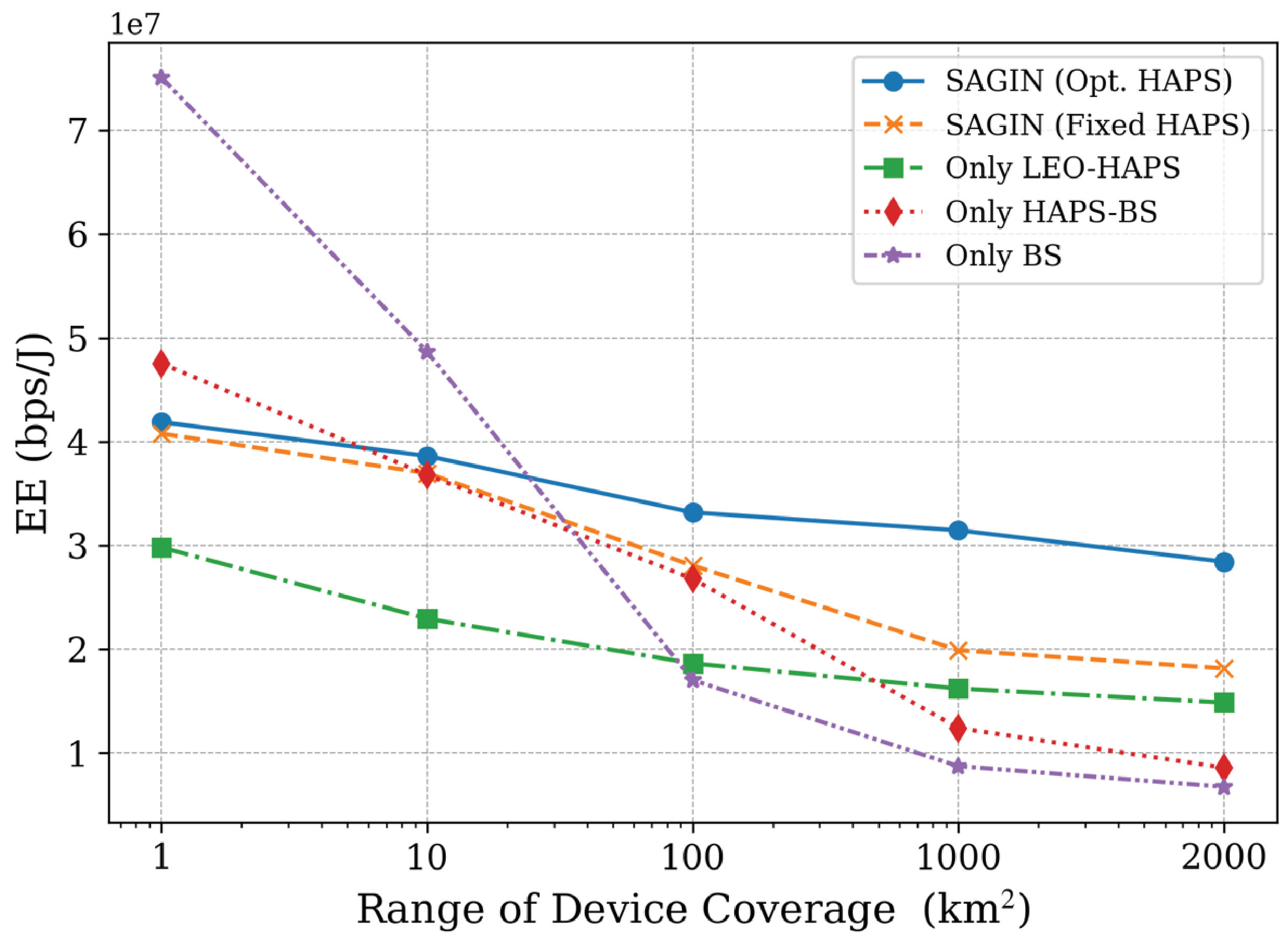}
\label{f.sim.6a}}
\subfigure[]
{\includegraphics[width=2.2in]{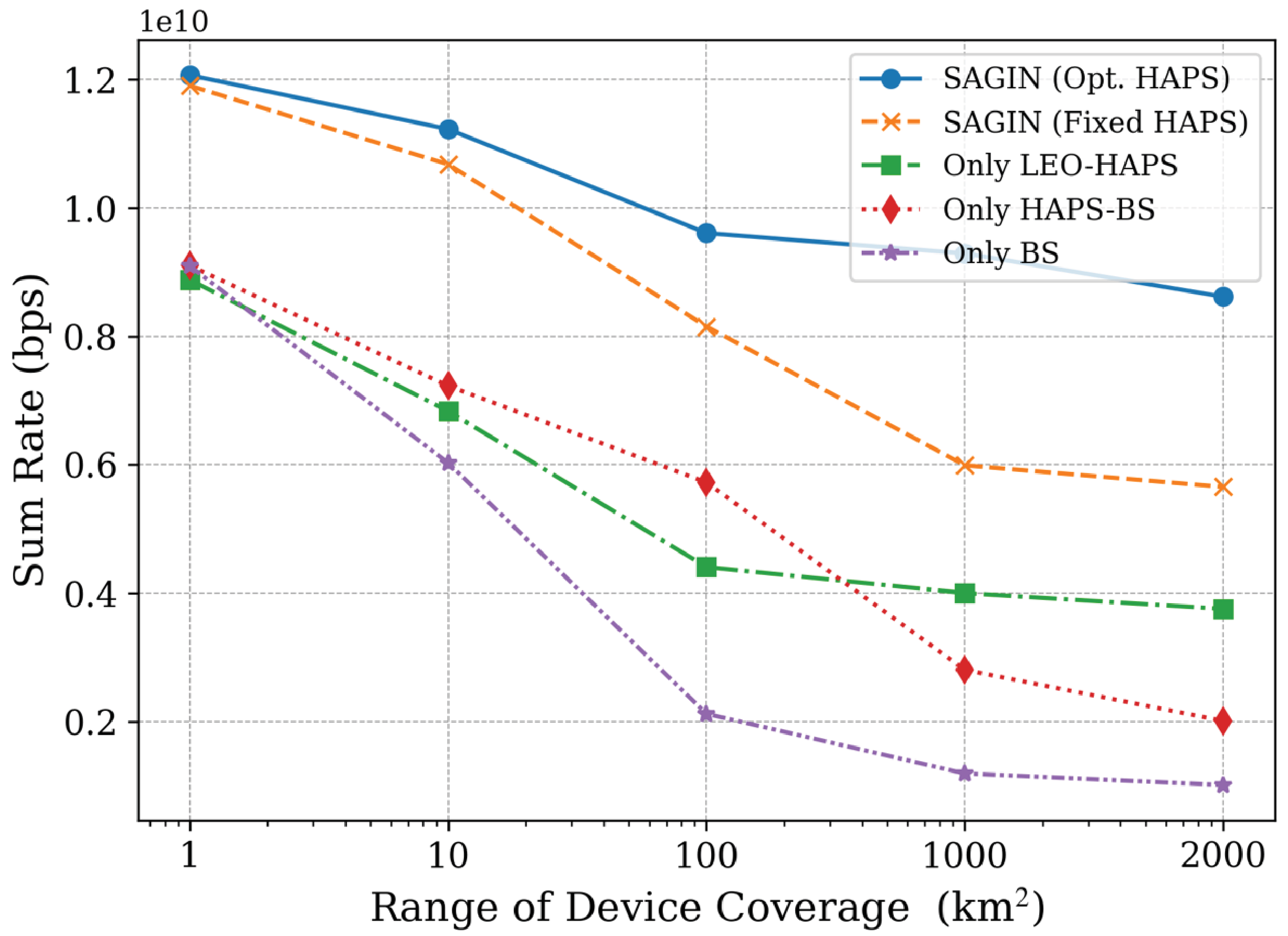}
\label{f.sim.6b}}
\subfigure[]
{\includegraphics[width=2.2in]{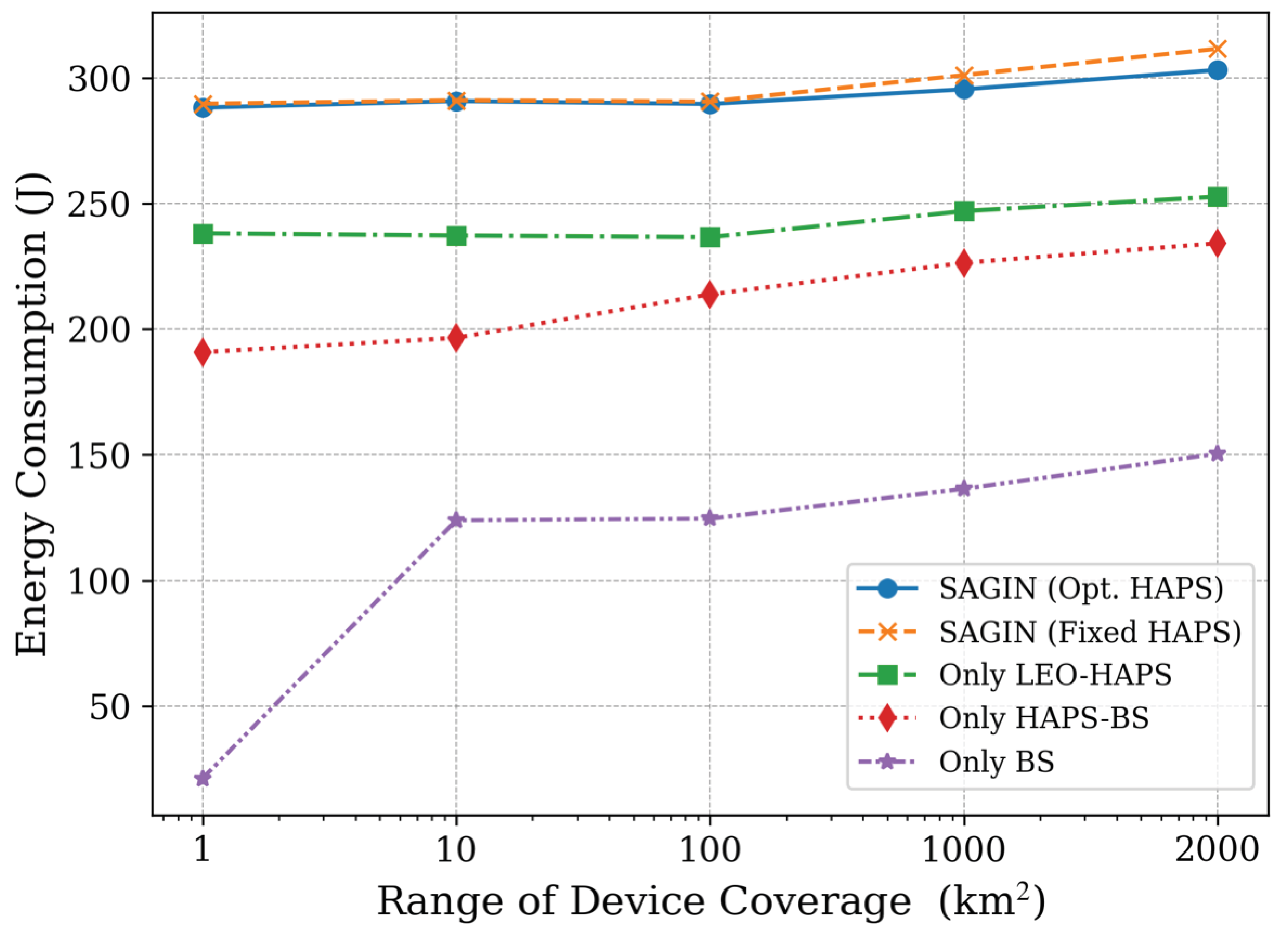}
\label{f.sim.6c}}
\caption{Performance of (a) EE, (b) sum rate and (c) power consumption versus different ranges of IoT device coverage area considering different SAGIN architectures.}
\label{f.sim.6}
\end{figure*}

Fig. \ref{f.sim.7} presents the EE performance versus different numbers of RIS elements under various deployment architectures: full MF-RIS deployment across all layers (SAGIN-MF-RIS), dual-layers (LEO-HAPS or HAPS-BS) and single-layer (either LEO, HAPS or BS) equipped with MF-RIS. It is important to note that while all node layers are present in the SAGIN architecture, only a subset of them are equipped with MF-RIS. Among all architectures, full-layer MF-RIS deployment consistently achieves the highest EE across all element numbers. The joint optimization of multi-layered MF-RIS supports diverse and resilient signal enhancement and interference suppression, leading to escalated gains in EE. In the dual-layer deployment, LEO-HAPS configuration outperforms HAPS-BS. This benefits arise primarily from the EH capabilities of MF-RIS elements deployed at LEO, allowing them to sustain signal enhancement even when solar energy is unavailable. While the HAPS-BS still benefits from additional reflective paths, its lack of RF-based EH support results in a comparatively lower EE than LEO-HAPS. For single-layer architectures, the BS-MF-RIS configuration achieves higher EE with fewer elements, owing to its shorter link distances and reduced signal attenuation compared to MF-RIS deployment at LEO or HAPS. Notably, the HAPS-only MF-RIS configuration exhibits the lowest EE among single-layer architectures, due to higher pathloss compared to BS-based setups and more constrained EH capabilities than those at LEO. In this context, the full-layer SAGIN-MF-RIS architecture leverages the combined advantages of spatial diversity, cooperative beamforming, and energy harvesting, significantly outperforming both dual-layer and single-layer counterparts.

\subsection{Different Architectures and Coverage}

Fig. \ref{f.sim.6} depicts the EE performance versus IoT device coverage area across different network architectures: SAGIN with full optimization, SAGIN with fixed HAPS deployment, LEO-HAPS, HAPS-BS, and the BS-only scenario. As the coverage area expands from 1 km$^2$ to 2000 km$^2$, all architectures exhibit a declining EE trend. In small-scale scenarios below 10 km$^2$, the BS-only configuration maintains relatively highest EE compared to SAGIN due to lower energy consumption, demonstrating its suitability for small-scale deployment such as the dense urban. However, as the coverage extends to hundreds or thousands of square kilometers, the EE of BS-only architecture deteriorates rapidly. Such decline is attributed to the extensive deployment of comparably high-power BSs to maintain coverage. Dual-layer architectures of LEO–BS and HAPS–BS offer partial flexibility but suffer from the potential coverage holes due to the absence of either HAPS or LEO nodes. As coverage increases, these limitations result in more explicit EE declination, particularly in the HAPS–BS case, where performance drops sharply beyond 100 km$^2$. Moreover, SAGIN with the fixed HAPS location lacks network reconfiguration capabilities, it still provides moderate EE performance. In contrast, the SAGIN-based architecture exhibits superior scalability and resilience in large-scale scenarios. Specifically, SAGIN with optimized HAPS deployment is capable of dynamically orchestrating resource across LEO satellites, HAPS, and ground stations. This architecture complements global-local coverage requirement while sustaining stable rate and alleviating the energy burden of long-range communications. Additionally, within ultra-wide coverage ranges above 1000 km$^2$, fully-optimized SAGIN demonstrates the slowest EE degradation, highlighting its resilience and adaptability. To this context, the results highlight a compelling trade-off between terrestrial-only deployment well-suited for small-scale regions and the heterogeneous multi-layered SAGIN architecture offering complementary coverage and dynamic coordination for wide-area services.

\subsection{Benchmark Comparison}

\begin{figure}[!t]
\centering
\includegraphics[width=3.3in]{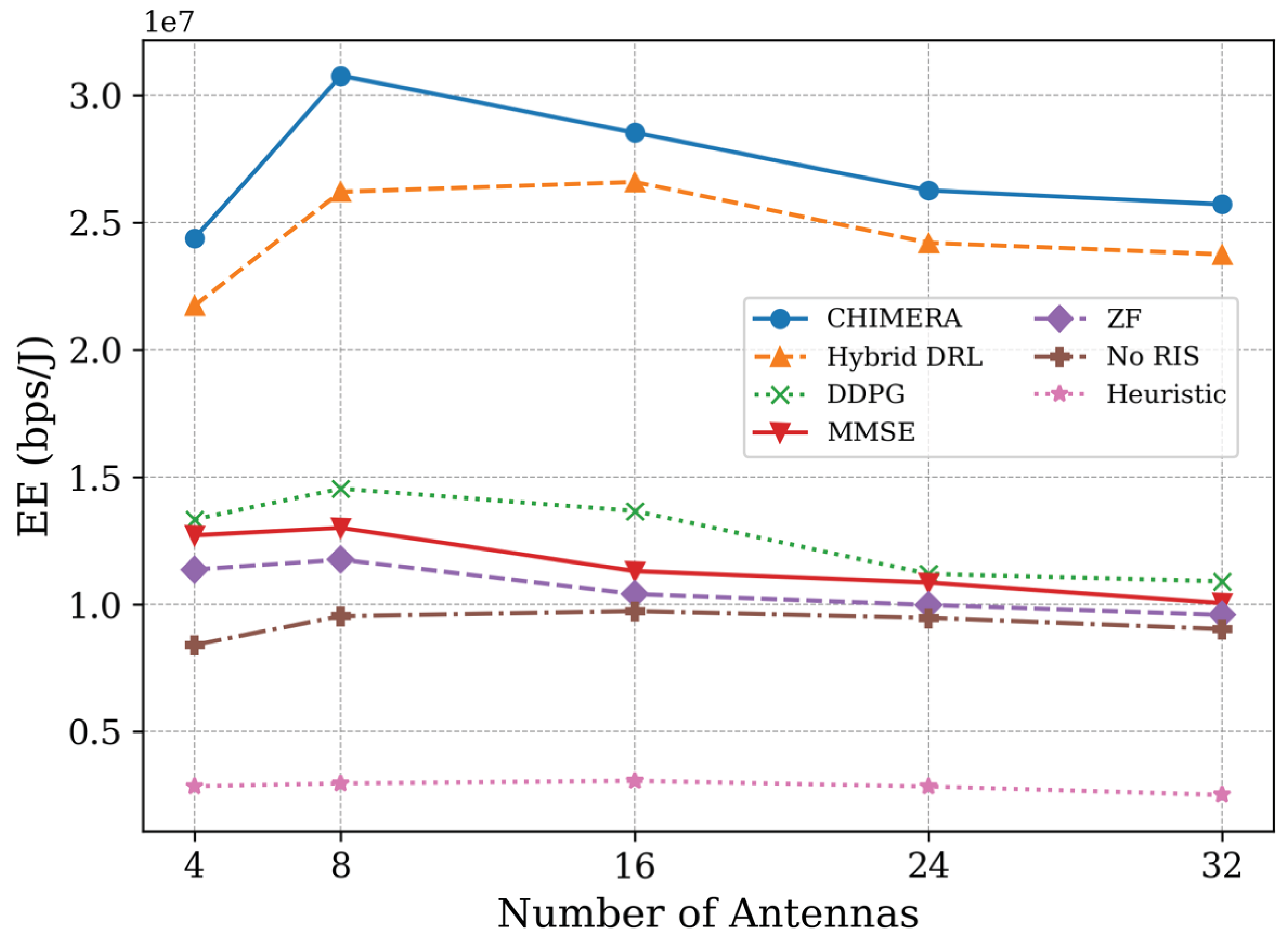}
\caption{Benchmark comparison of CHIMERA to the existing benchmarks under varying numbers of transmit antennas.} \label{f.sim.12}
\end{figure}

\begin{figure}[!t]
\centering
\includegraphics[width=3.3in]{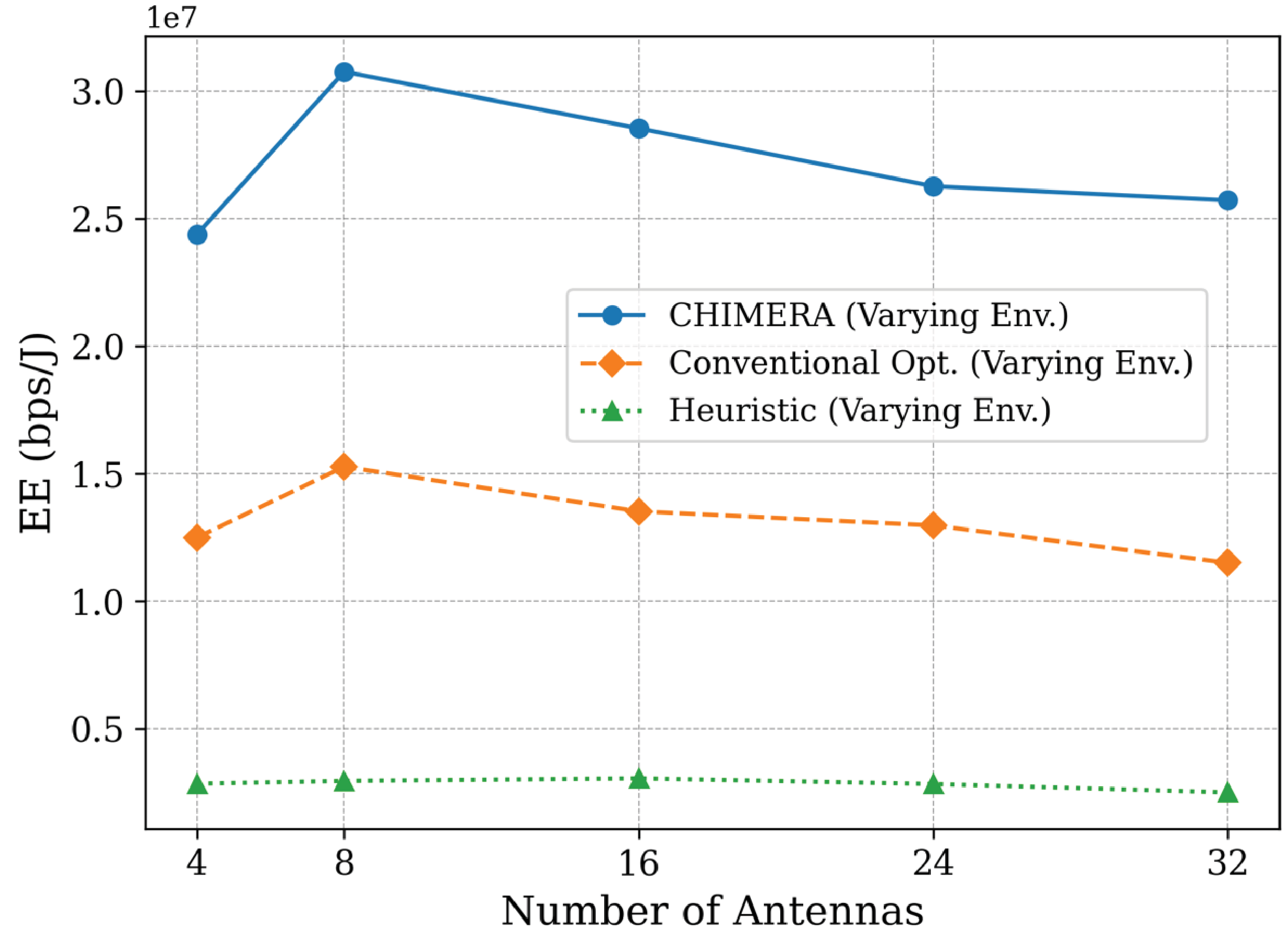}
\caption{EE performance of CHIMERA compared to heuristic and conventional optimization methods.} \label{f.sim.13}
\end{figure}

In Fig. \ref{f.sim.12}, we evaluate the EE performance of CHIMERA against several benchmark methods under varying number of transmit antennas, including hybrid DRL \cite{han}, DDPG \cite{my7}, conventional beamforming of minimum mean square error (MMSE) \cite{BM_MMSE} and zero-forcing (ZF) \cite{BM_ZF}, no-RIS deployment \cite{BM_noris}, and heuristic method using genetic algorithms (GAs) \cite{mobi}. Note that the GA executes bio-inspired operations of solution generation, elite selection, gene-style crossover and mutation under dynamic SAGIN environments. As the number of antennas increases from 4 to 8, all methods benefit from improved EE due to enhanced beamforming and spatial diversity. However, further increases up to 32 antennas lead to declining EE values, as the rising power consumption begins to outweigh the communication gains. The single-model hybrid DRL benchmark suffers from limited scalability and learning instability, resulting in a lower EE compared to CHIMERA. Without DQN assistance, pure DDPG struggles with discrete decision-making due to action quantization, yielding even lower EE than the hybrid DRL method. Traditional methods such as MMSE and ZF perform worse in EE, as they are unable to effectively handle complex interferences. The no-RIS case yields even lower EE, highlighting the importance of intelligent surface deployment. Intriguingly, the heuristic GA method performs worse than even the no-RIS scenario. This is due to its inefficiency under time-varying SAGIN channel conditions, where its reliance on evaluating large solution populations becomes impractical. In summary, the proposed CHIMERA framework consistently achieves the highest EE across all configurations, owing to its twin-model hybrid DRL structure combined with VAE-based semantic compression, enabling efficient learning in high-dimensional and dynamic environments. In Fig. \ref{f.sim.13}, we further compare the performance under varying SAGIN environments for the proposed CHIMERA, the conventional convex optimization approaches in the second footnote, and the heuristic GA method. It can be observed that the convex optimization methods achieve higher EE compared to GA, owing to the optimal solution derivation and sequential convex approximation process. However, the convex methods are not suitable for the highly dynamic SAGIN environment, as they lack instant adaptability and the obtained solution is typically reused across several LEO traversal slots. In contrast, the learning-based CHIMERA framework can perform instant inference, continuously interact with the environment, and deliver long-term optimal decisions.

\subsection{Computational Complexity Analysis}

\begin{table}[!t]
\centering
\scriptsize
\setstretch{1.2}
\caption{Computational Complexity}
\label{tab:complexity_a}
\begin{tabular}{|l|l|}
\hline
\textbf{Algorithm} & \textbf{Computational Complexity} \\
\hline
DQN & 
$\mathcal{O}\left( \sum_{c\in\mathcal{C}} N_c \cdot N_B N_L N_H^2 \right)$ \\
\hline
DDPG & 
$\mathcal{O}\left( \sum_{c\in\mathcal{C}} N_c \cdot 2 N_B N_L N_H^2 \right)$  \\
\hline
MADDPG & 
$\mathcal{O}\left( 2 N_B N_L N_H^2 \right)$ \\
\hline
VAE Compression & 
$\mathcal{O} \Big( 3 N_B ( \underbrace{ N_{L,\text{VE}} N_{H, \text{VE}}^2 + N_{L,\text{VD}} N_{H, \text{VD}}^2}_{\triangleq N_{\text{VAE}}} ) \Big)$ \\
\hline
\tabincell{l}{CHIMERA w/o \\ Twin-Model \& Compression} & $\mathcal{O}\left( 3 N_B N_L N_H^2 \right)$ \\
\hline
\tabincell{l}{CHIMERA \\ w/o Twin-Model} & 
$\mathcal{O}\left(	3 N_B (\upsilon N_L N_H^2 + N_{\text{VAE}})	\right)$ \\
\hline
CHIMERA w/o Compression & $\mathcal{O}\left(	6 N_B N_L N_H^2\right)$ \\
\hline
CHIMERA & $\mathcal{O}\left(	6 \upsilon N_B N_L N_H^2 + 3 N_B N_{\text{VAE}} \right)$ \\
\hline
\end{tabular}
\label{fig:performance_comparison}
\end{table}

Table~\ref{fig:performance_comparison} analyzes the computational complexity of the proposed CHIMERA scheme, its simplified version and the baseline methods. We define $N_B$ as the training mini-batch size, $N_L$ as the number of neural network layers and $N_H$ as the number of hidden neuron weights per layer. Note that for simplicity and for fair comparison, we use equivalent layers for each method and identical number of neurons for each layer. Notation of $0\leq \upsilon \leq 1$ is the compression ratio. Additionally, we denote $N_{\text{VAE}} = N_{L,\text{VE}} N_{H, \text{VE}}^2 + N_{L,\text{VD}} N_{H, \text{VD}}^2$ as the computational complexity of a single VAE compression model, where $N_{L,\text{VE/VD}}$ and $N_{H, \text{VE/VD}}$ indicate the number of layers and of neurons of the encoder/decoder, respectively. Moreover, we conduct several rounds and for averaging the per-step time overhead required. Simulations are performed using an Intel Core i7-13700 central processing unit (CPU) and NVIDIA GeForce RTX 4060 graphics processing unit (GPU). Each agent of 
DQN, 
DDPG, 
MADDPG, 
CHIMERA w/o twin-model \& compression, 
CHIMERA w/o twin-model,
CHIMERA w/o compression, and 
CHIMERA requires a storage memory of $\{2.79, 2.62, 1.57, 3.15, 4.12, 6.29, 7.3\}$ MB and the execution time of $\{23.1, 14.8, 3.71, 7.7, 4.5, 9.16, 6.96\}$ s per step. The results indicate that CHIMERA significantly reduces runtime compared to the non-compressed framework, demonstrating the practical benefit in accelerating training. Even though with the fastest execution time of MADDPG, the lack of flexibility in handling node specific actions limits their learning capability in large-scale complex systems. Note that the processing speed of deep learning techniques is relative, as it highly depends on the computing capability of the CPU/GPU and the degree of parallelism in the system architecture. For instance, second-level delays can be reduced to the millisecond level, thereby better matching the coherent channel time. In summary, the proposed CHIMERA framework effectively balances computational complexity and execution efficiency, which is especially well-suited for real-time operation in future joint communication and computing systems.

\section{Conclusions} \label{sec_con}
	We have proposed a SAGIN-MF-RIS architecture by deploying MF-RISs across LEO satellites, HAPS, and ground BSs to enable EE-based joint communication and computing optimization under stringent rate, power and latency constraints. To solve the high-dimensional discrete-continuous and non-convex problem, we have proposed a CHIMERA framework, incorporating semantic state-action compression using VAE and a hybrid DRL structure combining DQN and DDPG in a multi-agent manner. The twin-model and parametrized sharing mechanisms are designed to enhance learning stability, accelerate learning speed, and improve solution quality. Simulation results have demonstrated that CHIMERA substantially outperforms both centralized and decentralized DRL benchmarks and conventional methods. The results also highlight a compelling trade-off between terrestrial-only deployment well-suited for small-scale regions and the heterogeneous multi-layered SAGIN architecture for complementary coverage and dynamic coordination in wide-area services. Moreover, the proposed SAGIN-MF-RIS architecture achieves superior EE across a wide range of system configurations of fixed-EH, conventional RIS and no-RIS as well as of standalone network architectures of LEO satellites, HAPS or ground BSs.

\bibliographystyle{IEEEtran}
\bibliography{IEEEabrv}
\end{document}